\documentclass[letterpaper,journal]{IEEEtran}
\usepackage{amsmath,amsfonts}
\usepackage{algorithmic}
\usepackage{algorithm}
\usepackage{array}
\usepackage[caption=false,font=normalsize,labelfont=sf,textfont=sf]{subfig}
\usepackage{textcomp}
\usepackage{stfloats}
\usepackage{url}
\usepackage{verbatim}
\usepackage{graphicx}
\usepackage{cite}
\usepackage{amssymb}
\usepackage{multirow}
\usepackage{multicol}
\usepackage[table]{xcolor}
\usepackage{microtype}
\usepackage{multirow}
\usepackage{amsmath}
\usepackage{amssymb}
\usepackage{graphicx}
\usepackage{algorithmic}
\usepackage{array}
\usepackage{url}
\usepackage{booktabs}
\usepackage{microtype}
\usepackage{float}
\usepackage{bbm}
\usepackage{enumitem}
\usepackage{xcolor}
\usepackage{booktabs}
\usepackage{multirow}
\usepackage{graphicx}
\usepackage{pifont}
\usepackage{colortbl}
\usepackage{xcolor}
\usepackage{csquotes}
\usepackage{colortbl}
\newcommand{\cmark}{\ding{51}}

\usepackage[
    pagebackref=true,
    breaklinks=true,
    colorlinks=true,
    bookmarks=false
]{hyperref}
\usepackage{pifont}
\usepackage{todonotes}
\usepackage{makecell}
\begin{document}

\title{DeformStruct: Deformable Structure Modeling for Point-Supervised Temporal Action Localization}
% Unifying Query Reformation, Optimal Transport and DeformStruct in Detection Transformer for Point-Supervised Temporal Action Localization}
% From Points to SubActions: Enhancing Query Reformation and Optimal Transport with DeformStruct for PTAL}
% From Points to Sub-actions: Unifying Query Reformation, Structural Alignment, and Optimal Transport for PTAL
% DeformStruct for Point-supervised Temporal Action Localization}
% Boosting Point-Supervised Temporal Action Localization via Semantic Consensus and Structural Alignment}
% Boosting Point-Supervised Temporal Action Localization through Integrating Query Reformation， Optimal Transport and DeformStruct}
% From Points to Parts: Deformable Alignment for Structure-Aware Point-supervised Temporal Action Localization}
% Point-to-Instance Temporal Action Localization with Explicit Internal Structure Modeling}

\author{Mengnan~Liu,
        Sanping Zhou,~\IEEEmembership{Member,~IEEE,} 
        Kun Xia,
        Yonghao Dong,
        Jie Wang,~\IEEEmembership{Senior Member,~IEEE,} 
        % Nanning~Zheng,~\IEEEmembership{Fellow,~IEEE,}
        and~Gang~Hua,~\IEEEmembership{Fellow,~IEEE}
        % and~Le~Wang,~\IEEEmembership{Senior Member,~IEEE}
       
\thanks{This work was supported in part by the National Key Research and Development Project under Grant 2024YFB4708100, the Fundamental and Interdisciplinary Disciplines Breakthrough Plan of the Ministry of Education of China under Grant JYB2025XDXM504, National Natural Science Foundation of China under Grants 62572384, U24A20325, 62125305, 62503384, and Key Research and Development Plan of Shaanxi Province under Grant 2024PT-ZCK-80. Kun Xia was supported by NSFC Grant 62503384.~({\it Corresponding author: Sanping Zhou, E-mail: spzhou@mail.xjtu.edu.cn.})}
\thanks{M. Liu and S. Zhou are with the State Key Laboratory of Human-Machine Hybrid Augmented Intelligence, Institute of Artificial Intelligence and Robotics, Xi'an Jiaotong University, Xi'an, Shaanxi 710049, China. E-mail: Liumn@stu.xjtu.edu.cn,  spzhou@mail.xjtu.edu.cn. 

\IEEEcompsocthanksitem K. Xia is with the School of Computer Science and Technology, Xi'an Jiaotong University, Xi'an, Shaanxi 710049, China. E-mail: kunxia@xjtu.edu.cn.
\IEEEcompsocthanksitem Y. Dong is with the School of Software Engineering, Xi’an Jiaotong University, Xi’an 710049, China. E-mail: yhdong36@xjtu.edu.cn.
\IEEEcompsocthanksitem J. Wang is with the MoE Key Laboratory of Brain-Inspired Intelligent Perception and Cognition, University of Science and Technology of China, Hefei 230027, China. E-mail: jiewangx@ustc.edu.cn.
\IEEEcompsocthanksitem G. Hua is with Amazon Alexa AI, Bellevue, WA 98004 USA. E-mail: ganghua@gmail.com.
}}

\markboth{Journal of \LaTeX\ Class Files,~Vol.~14, No.~8, August~2021}%
{Shell \MakeLowercase{\textit{et al.}}: A Sample Article Using IEEEtran.cls for IEEE Journals}

% Remember, if you use this you must call \textit{ie}EEpubidadjcol in the second

\maketitle

\begin{abstract}

Point-supervised Temporal Action Localization (PTAL) aims to localize action instances using only a single annotated timestamp for each instance. Such sparse supervision offers limited guidance for modeling the structural organization of an action instance. Existing PTAL methods primarily infer action regions from classification responses, leaving the structural organization of multiple action-related parts within each instance insufficiently modeled. To address this problem, we propose DeformStruct, an instance-level deformable structure modeling framework for PTAL. DeformStruct progressively constructs the structure of each action instance through three stages. First, Structural Anchor Mining discovers multiple anchors from annotated points and action-responsive regions, providing structural nodes grounded in the current instance. Second, Deformable Part Alignment associates latent structural parts with these anchors through deformation-aware matching, allowing their temporal supports to adapt to instance-dependent variations. Third, Relational Structural Organization explicitly models the relative temporal relations among the aligned parts, forming a relational representation that characterizes how different action-related parts are organized together.
Extensive experiments on THUMOS14, ActivityNet-v1.3, GTEA, and BEOID demonstrate consistent improvements in point-supervised temporal localization and validate the effectiveness of deformable structural modeling.
\end{abstract}

\begin{IEEEkeywords}
Temporal action localization, Weakly-supervised learning, Video understanding.
\end{IEEEkeywords}

\section{Introduction}
\label{introduction}
 \begin{figure*}[t]
	\centering
	\includegraphics[width=\linewidth]{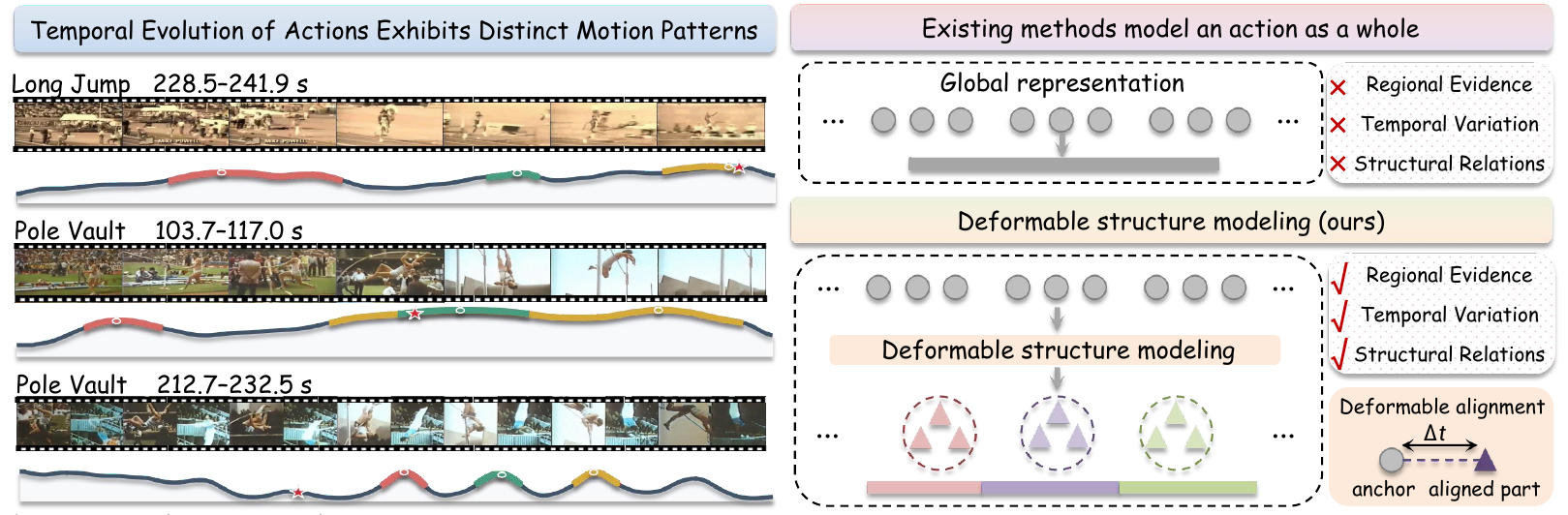}
\caption{
Motivation of deformable structural modeling for PTAL. A complete action instance may contain multiple temporally distributed action-related stages, whose locations, durations, and relative configurations vary across instances. Existing holistic modeling tends to overlook such regional evidence, temporal variation, and structural relations. Our DeformStruct instead models the action as a deformable structure composed of multiple action-related parts, adaptively aligning them to instance-specific temporal evidence and capturing their structural relations.}
	\label{fig:illustration}
\end{figure*}

\IEEEPARstart{T}{emporal} Action Localization~(TAL)~\cite{lin2019bmn, lin2018bsn, shi2022react} aims to detect actions of interest in untrimmed videos by identifying their temporal boundaries and action categories, which is a fundamental task in video understanding.
Fully-supervised TAL methods~\cite{shi2022react, lin2018bsn, lin2019bmn} have achieved remarkable performance with dense and precise temporal boundary annotations.
However, obtaining such annotations is time-consuming.
As a more annotation-efficient alternative, Weakly-Supervised Temporal Action Localization~(WTAL)~\cite{lee2020background} has been widely studied, where only video-level class labels are available.
Nevertheless, without instance-level temporal cues, WTAL methods often struggle to distinguish actions from background contexts.
To better balance annotation cost and localization performance, SF-Net~\cite{ma2020sf} introduces the Point-Supervised Temporal Action Localization~(PTAL) task, where each action instance is annotated with only a single timestamp.
% However, sparse labeling of individual frames leads to semantic ambiguity in determining action boundaries due to the lack of continuity in the highly sparse point-supervision scheme.

Existing PTAL methods~\cite{lee2021learning, zhang2024hr, xia2024realigning, liu2025stepwise} commonly follow a Multiple Instance Learning~(MIL)-based paradigm and derive localization cues mainly from frame- or snippet-level classification responses.
Although these methods have achieved promising performance, response-driven localization tends to emphasize locally discriminative regions, while point annotations provide supervision only at isolated temporal positions, leaving the broader temporal evolution of an action instance largely unconstrained.
As a result, existing approaches are effective at identifying salient action evidence, but remain limited in modeling how different action-related regions jointly form the structure of an action instance.
% This calls for a framework that can jointly reason over multiple action cues in the global temporal context.} Detection Transformer~(DETR)~\cite{carion2020end, zhu2020deformable} provides a promising basis for this purpose through global context modeling, query-based interaction, and direct set prediction.

% \textcolor{blue}{
% This limitation is particularly important because an action often evolves through multiple temporal stages exhibiting distinct motion patterns.
% A single point annotation captures only one local state of this evolution and therefore cannot sufficiently describe the structural pattern of the whole action instance.
% As illustrated in \figurename~\ref{fig:illustration}, our empirical analysis further reveals that
% (1) an action instance typically contains multiple distinct action-related regions,
% (2) their temporal configurations vary considerably across different instances, and
% (3) the relations among these regions provide complementary structural information beyond individual regions.
% These observations suggest that effective PTAL should model not only individual action parts, but also their instance-dependent organization and mutual relations.}

This limitation is particularly important because an action often evolves through multiple temporal stages exhibiting distinct motion patterns. For example, a long jump typically involves a run-up, take-off and landing, which jointly constitute the complete action instance. As illustrated in \figurename~\ref{fig:illustration}, these different stages are manifested as multiple action-related regions distributed over time. We further observe that the feature characteristics, temporal extents, and relative configurations of these regions vary considerably across instances. Specifically,  (1)~different action-related regions exhibit distinct semantic and motion characteristics; (2)~their temporal positions, durations, and relative intervals show substantial instance-dependent variations; and (3)~these regions, together with their relative temporal configurations, form an instance-specific temporal structure that can be obscured when an action is modeled as a single entity. These observations suggest that effective PTAL should go beyond locally discriminative responses and explicitly model deformable temporal structures that can adapt to instance-specific variations.

% \textcolor{blue}{
% This limitation is particularly important because an action often evolves through multiple temporal stages exhibiting distinct motion patterns. As illustrated in \figurename~\ref{fig:illustration}, a single point annotation provides only local supervision at one temporal position, whereas an action instance often contains multiple action-related regions distributed over time. We further observe that the feature characteristics, temporal extents, and relative configurations of these regions vary considerably across instances. Specifically, (1) an action instance often contains multiple action-related regions exhibiting distinct semantic and motion characteristics; (2) their temporal positions, durations, and relative intervals exhibit substantial instance-dependent variation; and (3) these regions together with their relative temporal configurations form an temporal structure, which can be obscured when an action is modeled as a single entity. These observations suggest that effective PTAL should go beyond relying on locally discriminative responses and explicitly model deformable temporal structures that can adapt to variations.}

To this end, we propose \textbf{DeformStruct}, a deformable structure modeling framework for point-supervised temporal action localization.
Instead of directly treating an action as a single entity, DeformStruct progressively constructs the structure of each action instance through three stages: Structural Anchor Mining~(SAM),  Deformable Part Alignment~(DPA), and Relational Structural Organization~(RSO).
The key idea is to first discover multiple action-related structural nodes from sparse supervision, then adapt their temporal supports to variations, and finally model their relative temporal relations.
In this way, DeformStruct captures both multiple action-related evidence regions and their relational structure.

Specifically, we first introduce \textbf{Structural Anchor Mining} to discover multiple anchors from sparse action evidence. Starting from the annotated point, we identify additional highly responsive temporal regions and apply temporal suppression to avoid repeatedly selecting neighboring responses.
The resulting anchors correspond to complementary action-related regions within the current instance, providing a sparse set of structural nodes for subsequent modeling. Compared with representing an action using only the annotated point, these anchors expose richer action evidence across the action instance.

Given these anchors, we further introduce \textbf{Deformable Part Alignment} to establish adaptive correspondences between latent structural parts and action anchors. Since instances of the same action category may vary considerably in duration, execution speed, and temporal composition, fixed temporal assignments are often inadequate. We therefore employ one-to-one deformation-aware matching to associate latent structural parts with the mined anchors, allowing their temporal supports to adapt to variations. This process produces a set of adaptively aligned parts that characterize the structural composition of each action instance.
Building on these aligned parts, we introduce \textbf{Relational
Structural Organization} to further model how they are
organized within the instance. RSO maps the aligned part
features into a shared structural coordinate space and derives
their pairwise temporal relations through coordinate differences.
By representing all parts within a common coordinate system,
RSO captures their instance-specific relative temporal organization.

As a result, our approach shows significant improvements compared to previous PTAL methods and achieves competitive performance with state-of-the-art methods on four popular benchmark datasets: THUMOS14~\cite{jiang2014thumos}, ActivityNet-v1.3~\cite{caba2015activitynet}, GTEA~\cite{lei2018temporal} and BEOID~\cite{damen2014you}.

To summarize, our contributions are as follows:
\begin{itemize}
    \item We introduce \textbf{DeformStruct}, an instance-level deformable structure modeling framework for point-supervised temporal action localization. It models an action instance as a set of structurally related parts, providing a new perspective for exploiting sparse point supervision beyond independent classification responses.

    \item We propose \textbf{Structural Anchor Mining} and \textbf{Deformable Part Alignment} to discover complementary action-related evidence and adaptively instantiate latent parts at instance-specific temporal locations, accommodating structural variations without relying on fixed temporal partitions.

  \item We introduce \textbf{Relational Structural Organization} to model the relative temporal relations among aligned parts, allowing the model to characterize both multiple action-related evidence regions and their relational structure.

    \item Extensive experiments on THUMOS14, ActivityNet-v1.3, GTEA, and BEOID demonstrate the effectiveness of DeformStruct. Detailed analyses further validate the benefits of anchor discovery, deformable alignment, and relational organization.
\end{itemize}

% \endgroup

We note that a conference version of this work appeared at CVPR 2025~\cite{liu2025boosting}. The conference version mainly exploited sparse point supervision through semantic query construction and optimal transport-based pseudo-segment supervision. Building upon this foundation, the present manuscript extends the original formulation from point-level semantic exploitation to deformable structure modeling, enabling the model to capture not only action-related evidence but also how multiple action-related parts are relatively organized within each action instance.
The major extensions over the conference version are summarized as follows:
\begin{itemize}

    \item We extend the modeling target from isolated point-level semantics to the structural organization of an entire action instance, explicitly considering multiple motion parts and their instance-dependent variations.

    \item We introduce a new structural modeling pipeline that discovers complementary action evidence, establishes adaptive part-to-evidence correspondences, and further captures the relations among these parts.

   % \item We augment the original framework with {relational structure modeling}, enabling the representation to capture the relative organization among different action-related components rather than modeling them independently.
   
    \item We substantially expand the experimental study with structural analyses, mechanism-level ablations, robustness evaluations, and efficiency comparisons to better understand the effectiveness of the extended framework.

\end{itemize}

This paper is organized as follows: Section~\ref{sec:related-work} briefly
reviews related work. We present the technical details of the proposed method in
Section~\ref{sec:methodology}. Experimental results and discussions are
presented in Section~\ref{sec:experiments-and-discussions}. Finally, we conclude
in Section~\ref{sec:conclusion}.

\section{Related Work}
\label{sec:related-work}

We briefly review related work in fully-supervised temporal action localization, weakly-supervised temporal action localization, point-supervised temporal action localization and part-based learning.

\noindent\textbf{Fully-Supervised Temporal Action Localization}
relies on accurately annotated temporal coordinates, assigning precise labels to each snippet during the training phase. This complete supervision enables high performance in both classification and localization tasks due to the availability of action boundaries. There are three primary approaches in this domain. (1) Proposal-based methods~\cite{chao2018rethinking,wang2022rcl,xu2017r,zhao2021video,xu2020g, su2023multi} generate a temporal window that marks the start and end snippets of an action. These methods then refine the boundaries through a regression head to enhance accuracy. (2) Proposal-free methods~\cite{xia2022dual,xia2022learning,xia2023exploring,lin2021learning,nag2022proposal, zeng2021graph} focus on predicting the likelihood of each snippet containing an action. By identifying the snippet with the highest probability, they determine the action boundaries without relying on pre-defined proposals. (3) Transformer-based~\cite{zhang2022actionformer,tan2021relaxed,liu2022end,cheng2022tallformer} methods utilize the Transformer architecture to facilitate action localization, achieving impressive results on various Temporal Action Localization benchmarks.
Despite the effectiveness of these approaches, the substantial effort required for annotating each snippet poses a significant challenge.

\noindent\textbf{Weakly-Supervised Temporal Action Localization} operates under a framework where only video-level labels are provided during training. This approach significantly reduces the annotation burden since frame-level labels are unnecessary. Existing methods can be broadly grouped into three categories.
(1)~The attention-based methods~\cite{nguyen2018weakly,nguyen2019weakly,zhai2020two,liu2019weakly} assess frame-level features to compute probabilities and utilize the resulting attention maps to guide the classification process.
(2)~The MIL-based methods~\cite{wang2017untrimmednets,paul2018w,liu2019completeness,zhang2019adversarial, zhu2023contextloc++, yang2021background, xu2023bilateral, liu2021weakly, chen2023uncertainty, yang2025cross, fu2023semantic} generate frame-level class scores, \textit{i.e.}, the class activation sequence (CAS). These scores are then aggregated to form a video-level label, allowing for a holistic assessment of the action.
(3)~The graph-based methods~\cite{rashid2020action,yang2022acgnet,ullah2023ad,zeng2019graph} model each video snippet as a node within a graph, where edges are weighted according to the similarity between snippets. In this framework, related snippets are drawn closer together in the feature space, while unrelated ones are pushed apart, enhancing the model's ability to differentiate actions.

\noindent\textbf{Point-Supervised Temporal Action Localization} serves as a bridge between fully supervised and weakly supervised methods by requiring only a single label for each action instance. 
Moltisanti \textit{et al.}~\cite{moltisanti2019action} first proposed point-supervised learning in temporal action localization. SF-Net~\cite{ma2020sf} expands the point annotation to adjacent snippets to obtain the pseudo-label for training. Ju \textit{et al.}~\cite{ju2021divide} propose a novel two-stage framework, which divides the entire video into multiple clips and sequentially conquers these clips. LACP~\cite{lee2021learning} utilizes a greedy algorithm to find the optimal sequence of actions and learns the completeness of the action for the optimal sequence. HR-Pro~\cite{zhang2024hr} proposes that the high reliability of point annotations can be propagated at both snippet and instance levels. RCTSI~\cite{xia2024realigning} presents a novel proposal-level plug-in framework to relearn the aligned confidence of proposals generated by the base locators. SMBD~\cite{liu2025stepwise} proposes to search for action boundaries to obtain complete action segments. 

However, these methods mainly focus on propagating point-level
responses or recovering temporal boundaries, while the internal
organization of multiple action-related regions within an action
instance remains insufficiently explored.

\noindent\textbf{Deformable Alignment Mechanisms.}
 Early works such as deformable template models~\cite{coughlan2000efficient} and manifold part-based models~\cite{cootes1998active} achieved remarkable success in handling appearance variance by designing meaningful part patterns in a bottom-up manner. In particular, Pictorial Structure Models~\cite{felzenszwalb2005pictorial} and the seminal Deformable Part Models~\cite{felzenszwalb2009object} explicitly captured the geometric arrangement of parts through a set of springs connecting part pairs, utilizing deformation costs to penalize spatial deviations from the root position. Recently, this structural paradigm has been reinvigorated within deep learning architectures. DCN~\cite{dai2017deformable} augmented convolutional kernels by learning spatial offsets to accommodate irregular object layouts. Deformable DETR~\cite{zhu2020deformable} integrated this concept into Vision Transformers to enhance feature representativeness.

In this study, we explore deformable structural modeling under point-supervised temporal action localization.
Unlike conventional part-based formulations that rely on predefined structures or stronger temporal supervision, our setting provides only a single annotated timestamp for each action instance.
We therefore use instance-specific evidence anchors as temporal references and adaptively instantiate latent parts through deformation-aware matching.
The resulting aligned parts are further organized through their relative temporal relations, enabling structural modeling without predefined partitions.

\begin{figure*}[t]
	\centering
	\includegraphics[width=\linewidth]{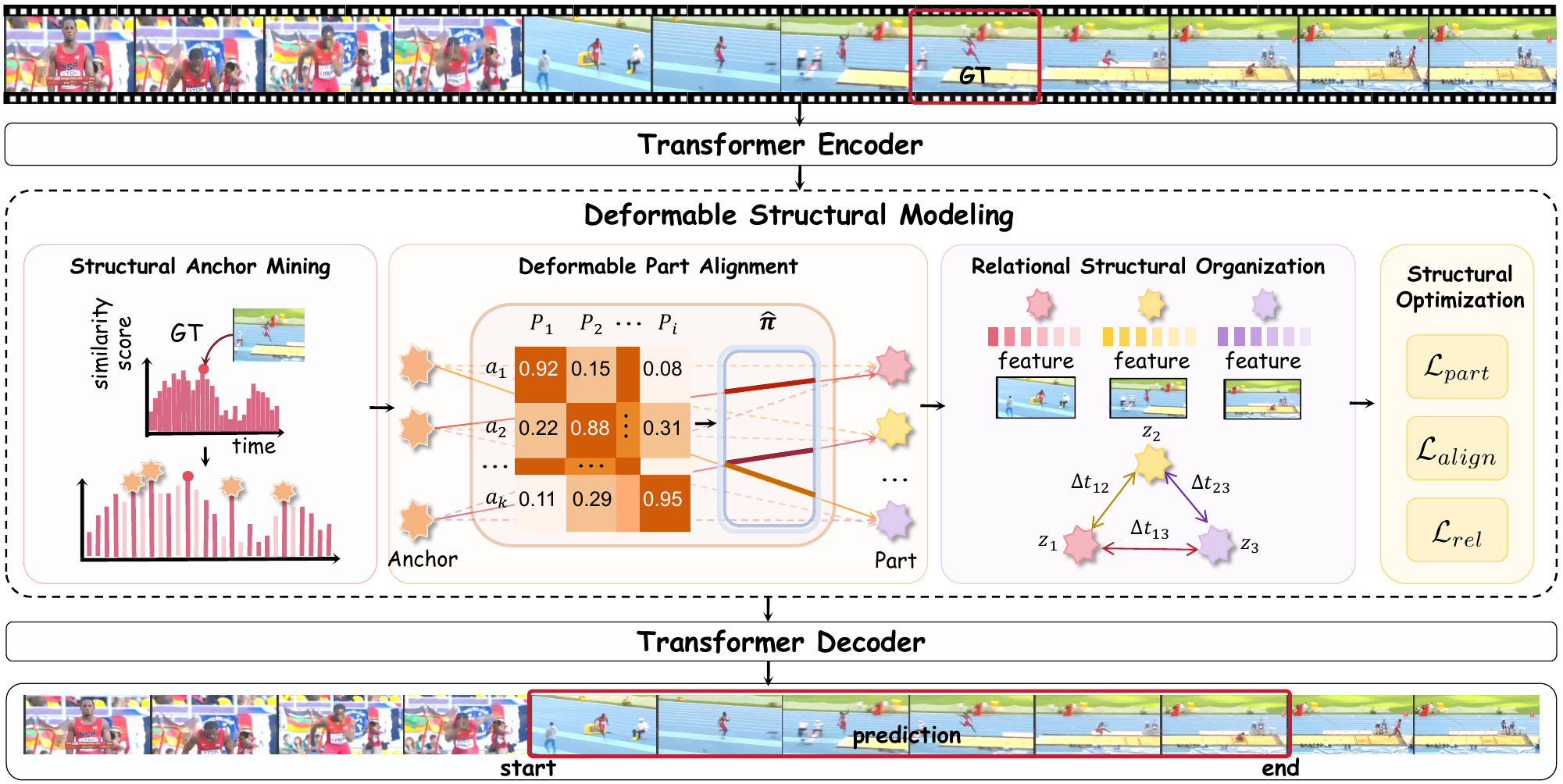}
	% \caption{ \textcolor{blue}{Overview of the proposed P2P-Align framework. First, pre-extracted video features enhanced with positional embeddings are processed by a Transformer Encoder to generate feature maps $X_{E}$. The framework then branches into three synergistic components to handle sparse supervision:
% (1) pseudo-label Generation: The classification head produces class activation scores (CAS), which are aligned with point labels via Optimal Transport (OT) to generate high-quality pseudo-labels (Start, End).
% (2) Deformable Part Alignment: This newly introduced module first mines part-level anchors using Iterative Temporal NMS. It then aligns latent parts with learnable deformation parameters to these anchors via bipartite matching, jointly optimizing classification responses and temporal deformation costs under the DPA loss to model the internal structural variability of actions.
% (3) Query Reformation: Point labels are encoded into semantic consensus queries to provide global priors for the Transformer Decoder.
% Finally, the decoder's prediction segments are optimized end-to-end using both the OT-generated pseudo-labels and the structural constraints from the Deformable part alignment module.}}

\caption{
Overview of the proposed \textbf{DeformStruct} framework.
Given pre-extracted video features and point annotations, a Transformer Encoder first produces the encoded feature sequence $\mathbf{X}_{E}$.
DeformStruct progressively models each action instance through three stages: \textit{Structural Anchor Mining} discovers multiple action-related anchors, \textit{Deformable Part Alignment} adaptively aligns latent parts with these anchors to accommodate temporal variations, and \textit{Relational Structural Organization} models the relative temporal relations among the aligned parts to characterize their structural configuration.
Finally, the Transformer Decoder predicts candidate action segments, and the framework is jointly optimized with localization and structural objectives.}
	\label{fig:framework}
\end{figure*}

\section{Proposed Method}
\label{sec:methodology}

\subsection{Preliminary}
\noindent\textbf{Problem Definition}. Point-supervised temporal action localization~(PTAL) aims to accurately retrieve the start and end frames of the action of interest in an untrimmed video with only a single frame annotation per action. Given a labeled video, each action is annotated with a timestamp $t$ and its corresponding action category $y$. The PTAL model outputs predictions for each action, represented as $(\hat{s}, \hat{e}, \hat{y}, \hat{p})$, where $\hat{s}$ and $\hat{e}$ are the predicted start and end times, $\hat{y}$ is the predicted action category, and $\hat{p}$ denotes its confidence score.

\noindent\textbf{Feature Embedding}. Following conventions~\cite{lee2021learning, zhang2024hr}, we extract snippet-level features from consecutive fixed-length frames using a pre-trained model~(\textit{i.e.}, I3D~\cite{carreira2017quo}). The feature representation for each video is formulated as $\boldsymbol{X} \in \mathbb{R}^{T \times D}$, where $T$ is the number of snippets and $D$ is the feature dimension. 
The Transformer encoder maps the input features $\mathbf{X}$ to
context-enhanced representations
$\mathbf{X}_E\in\mathbb{R}^{T\times D}$.

\noindent\textbf{Overview}.
We build upon our conference version~\cite{liu2025boosting} and
introduce DeformStruct as a structural supervision branch on top of
the encoded video representation $\mathbf{X}_E$. Given $\mathbf{X}_E$, DeformStruct progressively encourages the encoder to learn structure-aware representations through \textit{Structural Anchor Mining}, \textit{Deformable Part Alignment}, and \textit{Relational Structural Organization}.
The resulting structural modeling complements the original localization framework by capturing multiple action-related parts, adapting them to temporal variations, and modeling their relative temporal relations.
For training supervision, we retain the pseudo-label generation mechanism from our conference version, where the generated pseudo segments are used to supervise the localization decoder. The overall framework is illustrated in Fig.~\ref{fig:framework}.
% \noindent\textbf{Overview}.  
% \textcolor{blue}{We employ the DETR architecture as the baseline for point-supervised temporal action localization and focus on modeling the structure of actions under sparse supervision.  To this end, we introduce \textbf{DeformStruct}, which progressively constructs an deformable structure from three aspects: (1) \textbf{Structural Anchor Mining} discovers multiple complementary action-related regions as structural nodes; (2) \textbf{Deformable Part Alignment} adaptively associates latent structural parts with these anchors through deformation-aware matching; and (3) \textbf{Relational Structural Organization} models the pairwise relations among the aligned parts to characterize their structural organization within each action instance. 
% The overall pipeline is illustrated in Fig.~\ref{fig:framework}. We elaborate on each part in the following.}

\subsection{DeformStruct}

Point supervision provides only sparse local cues for each action instance, making it difficult to characterize its complete structure.
In practice, an action instance often evolves through multiple temporal stages with distinct motion patterns. For example, a long jump typically involves the run-up, take-off, and landing stages, which jointly constitute the complete action instance. The temporal locations, durations, and relative configurations of these stages may vary considerably across different instances.
To explicitly capture such structural variations, we propose \textbf{DeformStruct}, which models each action instance as a deformable composition of latent parts.

DeformStruct progressively constructs the structure through three stages: \textit{Structural Anchor Mining}, \textit{Deformable Part Alignment}, and \textit{Relational Structural Organization}.
Structural Anchor Mining discovers multiple action-related anchors from sparse point supervision to provide structural evidence.
Deformable Part Alignment adaptively associates latent parts with these anchors, allowing their temporal supports to accommodate variations across action instances.
Relational Structural Organization further models the relative temporal relations among the aligned parts to characterize their structural configuration.
The resulting structure is jointly optimized through the structural consistency objectives described below. 
These structural components jointly regularize the shared encoder
representation, enabling the localization branch to benefit from
instance-level structural cues.
 
% While semantic consensus queries provide global class-level priors, they inherently fail to capture the fine-grained structural evolution within individual action instances. A complete action typically consists of multiple temporal stages, such as the run-up, take-off, and landing phases of a long jump. To explicitly model this internal compositionality, we propose a DeformStruct framework. This framework comprises three pivotal steps: Structural Anchor Mining, Latent Part Matching, and Structural Consistency Optimization. Through this procedure, each action instance is represented as a flexible composition of deformable latent parts, which are dynamically grounded to temporally distributed part-level anchors. These parts jointly capture the temporally distributed internal composition of the action.

\subsubsection{Structural Anchor Mining}
\label{anchor mining}
Traditional point-supervised paradigms are often constrained by \enquote{Single-frame View Bias}, where the feature at the sparse annotation point is treated as the sole training sample. Relying solely on such transient temporal snapshots inevitably overlooks diverse action-relevant patterns distributed along the temporal dimension, which is essential for comprehensively characterizing the internal temporal composition of the action.

To rectify this limitation, we propose Structural Anchor Mining.
Operating via a Query-and-Expand mechanism, we leverage the annotated
point as a Semantic Query to explore the temporal interval bounded by its preceding and succeeding annotations, identifying latent regions that exhibit high semantic consistency with the annotated action.
This effectively leverages the internal statistics of the video itself to expand the \enquote{single view} into a comprehensive set of candidate anchors. To prevent anchor redundancy (\textit{i.e.}, multiple anchors collapsing onto the same discriminative region), we further incorporate an iterative temporal NMS~\cite{neubeck2006efficient} mechanism.
Specifically, let $\{t_g\}_{g=1}^{G}$ denote the temporally ordered point annotations in a video. For the current annotated point $t_g$,
we extract its feature as the instance prototype $\mathbf{p}_{\mathrm{inst}}=\mathbf{X}_{E,t_g}$. To reduce interference from other action instances, anchor mining is restricted to the temporal interval bounded by its preceding and succeeding annotations: $\mathcal{R}_g=[t_{g-1},t_{g+1}]$. For the first and last annotations, the corresponding video boundary is used when one neighboring annotation is unavailable.

To facilitate global semantic interaction between the prototype and the video representation, we construct a Semantic Resonance Map $\mathbf{S}_{r} \in \mathbb{R}^{T}$ by measuring the dot-product similarity between the prototype and the video feature sequence,
\begin{equation}
\label{eq:similarity}
\mathbf{S}_{r} = \mathbf{p}_{inst} \cdot \mathbf{X}_E^\top.
\end{equation}
High activation values in $\mathbf{S}_{r}$ indicate that the snippets are semantically resonant with the annotated core. However, simply selecting the top-$K_a$ values often yields redundant anchors clustered around the most discriminative frames, failing to cover temporally distributed action-related evidence. To capture temporally diverse action-related evidence, we employ an NMS strategy to mine a temporally diverse candidate set $\mathcal{A}$.

Formally, we initialize $\mathbf{S}_{r}^{(0)} = \mathbf{S}_{r}$ and denote the number of candidate anchors by $K_a$. The mining process is conducted iteratively for $k=1, \dots, K_a$. In each step, we identify the timestamp $t_k$ with the maximum response and subsequently suppress its temporal neighborhood to encourage diversity,
\begin{align}
\label{eq:mining_nms}
t_k &= \arg\max_{t\in\mathcal{R}_g} \mathbf{S}_r^{(k-1)}(t), \\
\label{eq:suppression}
\mathbf{S}_{r}^{(k)}(t) &= 
\begin{cases} 
-\infty & \text{if } |t - t_k| \leq \delta \times T, \\
\mathbf{S}_{r}^{(k-1)}(t) & \text{otherwise},
\end{cases}
\end{align}
where $\delta$ is a predefined suppression radius. By leveraging the internal self-similarity of the video, this strategy effectively mines semantically responsive anchors that span various temporal regions, preventing them from collapsing into a single action phase. These anchors serve as a solid foundation for subsequent deformable part alignment (Sec.~\ref{part alignment}), enabling the model to perceive the action holistically rather than locally.

\subsubsection{Deformable Part Alignment}
\label{part alignment}

Given the set of candidate anchors $\mathcal{A} = \{\mathbf{a}_1, \dots, \mathbf{a}_{K_a}\}$ (as described in Sec.~\ref{anchor mining}), where each anchor $\mathbf{a}_{k} = \mathbf{X}_{E, t_k}$ represents the feature vector extracted at the mined timestamp $t_k$, we aim to model the instance-specific temporal structure of the action. Specifically, we formulate the action instance as a composition of $N_{part}$ latent parts, which are then optimally aligned with the mined anchors under deformable geometric constraints.

% \textbf{Parametric Part Scoring.}
We employ a set of learnable part-specific classifiers to produce response sequences for different latent parts. For the $i$-th latent part, we utilize a dedicated linear projection $\phi_i(\cdot)$ to generate the raw response sequence $\mathbf{S}_i \in \mathbb{R}^{T}$,
\begin{equation}
\label{eq:part_logit}
\mathbf{S}_i(t) = \phi_i(\mathbf{X}_{E, t}) = \mathbf{w}_i^\top \mathbf{X}_{E, t} + b_i,
\end{equation}
where $\mathbf{w}_i$ and $b_i$ are the learnable weight and bias for the $i$-th part classifier. The activation scores are then normalized via a sigmoid function $\sigma(\cdot)$.

% \textbf{Deformation-Aware Alignment.}
To model the temporal deformation of actions, we introduce a deformation cost that penalizes parts for deviating from their reference anchors. For the $i$-th part and a candidate anchor $\mathbf{a}_k \in \mathcal{A}$, the deformation penalty $\Psi_i$ at time $t$ is defined as a quadratic function of the temporal distance $d = |t - t_k|/T$,
\begin{equation}
\label{eq:deform_cost}
\Psi_i(t, t_k) = \gamma_{i,1} \cdot d + \gamma_{i,2} \cdot d^2,
\end{equation}
where $\gamma_{i,1}$ and $\gamma_{i,2}$ are non-negative learnable stiffness parameters for the $i$-th part, parameterized as
$\gamma_{i,r}=\operatorname{Softplus}(\tilde{\gamma}_{i,r})$, $r\in\{1,2\}$.
This allows the model to adaptively control the penalty for temporal deviations from the matched anchor.

We then determine the alignment score $V_{i, k}$ for assigning the $i$-th part to the $k$-th anchor by maximizing the deformation-aware objective over the temporal dimension,

\begin{equation}
\label{eq:latent_alignment}
V_{i,k} = \max_{t\in\mathcal{R}_g} \left( \sigma(\mathbf{S}_i(t)) - \lambda_{deform} \cdot \Psi_i(t, t_k) \right).
\end{equation}
Here, $\lambda_{deform}$ is a hyperparameter controlling the trade-off between classification score and temporal deformation penalty. This operation effectively searches for the optimal temporal position $t^*$ that balances semantic confidence and geometric consistency relative to the anchor.

% \textbf{Structured Bipartite Matching.}
Finally, to resolve ambiguities where multiple parts might compete for the same anchor, we formulate the assignment as a maximum weight bipartite matching problem. The optimal assignment $\hat{\pi}$ that maps each latent part to a unique anchor is obtained via the Hungarian algorithm,
\begin{equation}
\label{eq:hungarian}
\hat{\pi} =  \operatorname*{arg\,max}_{\pi} \sum_{i=1}^{N_{part}} V_{i, \pi(i)} .
\end{equation}
The resulting assignment grounds the abstract latent parts onto concrete temporal regions according to their alignment quality. 

% Notably, the matching does not require the latent-part indices to follow a predefined monotonic order, such as enforcing \(part_1<part_2<part_3\).
% Instead, the internal temporal structure is characterized by the temporal locations, coverage, and deformation of the matched parts.
% The resulting permutation $\hat{\pi}$ effectively grounds the abstract action parts onto the concrete video timeline, establishing a correspondence driven by the alignment quality.

\subsubsection{Relational Structural Organization}
\label{relational organization}

Although deformable matching adaptively associates individual latent parts with temporal regions, these aligned parts are still modeled independently. An action instance is jointly characterized by multiple motion parts, whose relative temporal configuration provides complementary information beyond the evidence of each individual part.
Therefore, we introduce Relational Structural Organization~(RSO) to explicitly characterize the relative organization among the deformably aligned parts.

After Hungarian matching, the $i$-th latent part is assigned to an anchor and obtains its optimal response position $t_i^*$ through deformation-aware matching.
We extract the encoder feature at this aligned position as,
\begin{equation}
\mathbf{z}_i
=
\mathbf{X}_{E,t_i^*},
\end{equation}
where $\mathbf{X}_E\in\mathbb{R}^{T\times D}$ denotes the encoded video feature sequence, $T$ denotes the temporal length, and $D$ denotes the feature dimension.
Accordingly, $\mathbf{z}_i\in\mathbb{R}^{D}$ represents the motion evidence captured by the $i$-th aligned part.

Directly predicting the relation of every part pair independently may produce mutually inconsistent pairwise relations.
To obtain a unified relational representation, we instead map all aligned parts into a shared one-dimensional structural coordinate space.
Specifically, the structural coordinate of the $i$-th part is formulated as,
\begin{equation}
q_i
=
\sigma
\left(
g(\mathbf{z}_i)
\right),
\end{equation}
where $g:\mathbb{R}^{D}\rightarrow\mathbb{R}$ denotes a lightweight two-layer MLP shared by all latent parts, and $\sigma(\cdot)$ denotes the sigmoid function.
The resulting $q_i\in[0,1]$ serves as a latent structural coordinate that characterizes the relative position of the $i$-th part in the learned structural space.
Importantly, the same mapping $g(\cdot)$ is applied to all parts, so that their structural coordinates are represented in a common coordinate system.

To characterize the temporal configuration of the aligned parts, we use their deformation-adjusted temporal positions as the structural reference. Specifically, the normalized temporal coordinate of the \(i\)-th aligned part is defined as,
\begin{equation}
p_i
=
\frac{t_i^*}{T},
\end{equation}
where \(t_i^*\) denotes the optimal temporal position obtained for the \(i\)-th latent part through deformation-aware alignment. The normalization reduces the influence of different video durations and provides a comparable temporal coordinate across action instances.

For each part pair $(i,j)$, we define the observed relative temporal relation as,
\begin{equation}
\label{delta}
\Delta_{ij}
=
p_j-p_i
=
\frac{t_j^*-t_i^*}{T},
\end{equation}
where the sign of \(\Delta_{ij}\) indicates the relative temporal order between the two aligned parts, while its magnitude characterizes their normalized temporal separation. Since \(t_i^*\) and \(t_j^*\) are obtained through instance-specific deformable alignment, \(\Delta_{ij}\) directly describes the relative temporal configuration of the instantiated parts without assuming a predefined ordering or fixed temporal partition.

Correspondingly, the relation predicted from the learned structural coordinates is defined as,
\begin{equation}
\widehat{\Delta}_{ij}
=
q_j-q_i.
\end{equation}
Rather than independently regressing every pairwise relation, this formulation derives all relations from the shared structural coordinates $\{q_i\}_{i=1}^{N_{part}}$.
Therefore, the predicted relations naturally satisfy the antisymmetric property,
\begin{equation}
\widehat{\Delta}_{ij}
=
-\widehat{\Delta}_{ji},
\end{equation}
and the transitive consistency,
\begin{equation}
\widehat{\Delta}_{ij}
+
\widehat{\Delta}_{jk}
=
\widehat{\Delta}_{ik}.
\end{equation}
These properties encourage different pairwise relations to form a coherent structural configuration.
In this way, RSO maps the deformably aligned parts into a shared latent structural coordinate space and characterizes their organization through relative coordinate differences.
Consequently, DPA determines \emph{where} individual motion parts are grounded, while RSO further describes \emph{how} these parts are organized relative to one another.

% The relational objective is formulated as,
% \begin{equation}
% \mathcal{L}_{rel}
% =
% \frac{1}{N_{part}(N_{part}-1)}
% \sum_{i\neq j}
% \operatorname{SmoothL1}
% \left(
% \hat{\Delta}_{ij},
% \Delta_{ij}
% \right),
% \end{equation}
% where $N_{part}$ denotes the number of latent parts.
% This objective explicitly encourages the aligned parts to preserve
% their relative configuration rather than being optimized
% independently.

% The final structural objective becomes,
% \begin{equation}
% \mathcal{L}_{struct}^{+}
% =
% \mathcal{L}_{struct}
% +
% \lambda_{rel}\mathcal{L}_{rel},
% \end{equation}
% where $\lambda_{rel}$ controls the contribution of relational structure
% modeling.
% In this way, DPA identifies and aligns individual action parts, while
% RSO further captures how these parts are relatively organized within
% each action instance.

\subsubsection{Structural Consistency Optimization}

To jointly optimize the discovered action evidence and the deformably aligned parts, we introduce three complementary structural objectives. The part classification loss promotes part-specific discrimination, the anchor-prototype alignment loss preserves the semantic reliability of the mined anchors, and the relational consistency loss regularizes the relative temporal organization of the parts.

First, we enforce a Part Classification Loss to encourage each latent part to establish a distinctive correspondence with its matched structural anchor. Let $\hat{\pi}$ denote the optimal assignment obtained by bipartite matching, where $\hat{\pi}(i)$ indicates the anchor assigned to the $i$-th latent part. Since $V_{i,k}$ jointly measures the part response and temporal deformation with respect to the $k$-th anchor, we directly optimize the deformation-aware matching scores as follows,
\begin{equation}
\label{eq:loss_cls}
\mathcal{L}_{\mathrm{part}}
=
\frac{1}{N_{\mathrm{part}}}
\sum_{i=1}^{N_{\mathrm{part}}}
\mathrm{CE}
\left(
\left[
V_{1,\hat{\pi}(i)}, \ldots,
V_{N_{\mathrm{part}},\hat{\pi}(i)}
\right],
i
\right).
\end{equation}
This objective encourages the anchor assigned to the $i$-th part to exhibit higher deformation-aware compatibility with that part than with the other parts. During back-propagation, the discrete assignment $\hat{\pi}$ is treated as fixed, while the matching scores $V_{i,k}$ remain differentiable. Therefore, $\mathcal{L}_{\mathrm{part}}$ jointly optimizes the part classifiers and the learnable deformation stiffness.

% This objective encourages each aligned position to be preferentially associated with its corresponding latent part, thereby promoting discriminative and part-specific structural responses.
Simultaneously, to guarantee that the selected anchors are semantically grounded to the action instance, we introduce an Anchor-Prototype Alignment Constraint. We minimize the distance between the prototype and the features of the anchors assigned to the latent parts. Formally, let $\mathbf{x}_{anchor}^{(i)}$ denote the feature vector of the anchor matched to the $i$-th part. The alignment loss is formulated as follows,
\begin{equation}
\label{eq:loss_align}
% \mathcal{L}_{align} = \frac{1}{N_{part}} \sum_{j=1}^{N_{part}} | \mathbf{p}_{inst} - \mathbf{x}_{anchor}^{(n)} |^2
% \end{equation}
\mathcal{L}_{align} = \frac{1}{N_{part}} \sum_{i=1}^{N_{part}} \| \mathbf{p}_{inst} - \mathbf{x}_{anchor}^{(i)} \|_{2}^{2}.
\end{equation}
This term encourages the mined anchors to remain semantically associated with the annotated action instance, providing reliable action-related evidence for deformable part alignment.

Finally, we impose a relational consistency constraint on the aligned parts. For each part pair $(i,j)$, the relative temporal relation $\Delta_{ij}$ defined in Eq.~(\ref{delta}) is used as the structural supervision.
Given the relation prediction $\widehat{\Delta}_{ij}$ produced by RSO, we define the relational consistency loss as,
\begin{equation}
\mathcal{L}_{\mathrm{rel}}
=
\frac{1}
{N_{\mathrm{part}}(N_{\mathrm{part}}-1)}
\sum_{i\neq j}
\mathrm{SmoothL1}
\left(
\widehat{\Delta}_{ij},
\Delta_{ij}
\right).
\end{equation}
By constructing the relational target directly from the deformable part positions, 
$\mathcal{L}_{\mathrm{rel}}$ encourages the learned structural coordinates to preserve the instance-specific organization of the aligned parts.

The final structural objective is formulated as,
\begin{equation}
\label{eq:loss_total}
\mathcal{L}_{struct}
=
\lambda_{part}\mathcal{L}_{part}
+
\lambda_{align}\mathcal{L}_{align}
+
\lambda_{rel}\mathcal{L}_{rel},
\end{equation}
where $\lambda_{part}$, $\lambda_{align}$, and $\lambda_{rel}$ are the
weighting parameters for part classification, anchor alignment, and
relational consistency, respectively.
% The final training objective for the DeformStruct module is as follows,
% \begin{equation}
% \label{eq:loss_total}
% \mathcal{L}_{struct} = \lambda_{part}\mathcal{L}_{part} + \lambda_{align} \mathcal{L}_{align},
% \end{equation}
% where $\lambda_{part}$ and $\lambda_{align}$ are the weighting parameters to balance the losses, which are set to 1 and 0.5, respectively.

% \endgroup

\subsection{Query Reformation}

Following our conference version~\cite{liu2025boosting}, we retain the Query Reformation mechanism to provide semantic priors for decoder initialization under point supervision.
Specifically, point-labeled features from all training videos are collected according to their action categories and clustered to obtain a set of semantic consensus prototypes.
These prototypes are then projected into decoder queries, providing class-aware semantic cues that guide the localization process.
Since the prototypes are constructed from sparse point annotations, this module mainly captures point-level semantic information and serves as the semantic initialization for the localization decoder.

\subsection{Pseudo-label Generation via Optimal Transport}

Following our conference version~\cite{liu2025boosting}, we employ an Optimal Transport~(OT)-based pseudo-label generator to provide segment-level supervision under point annotations.
Specifically, annotated action points and mined background points are treated as suppliers, while unlabeled snippets serve as consumers.
The transportation cost between supplier $i$ and consumer $j$ jointly considers semantic compatibility and temporal distance:
\begin{equation}
c(i,j)
=
d^{s}_{i,j}
+
\lambda_{OT}\frac{d^{p}_{i,j}}{T},
\end{equation}
where
$d^{s}_{i,j}=|p_i^{s}(y_i)-p_j^{s}(y_i)|$
measures their semantic discrepancy,
$d^{p}_{i,j}=|t(i)-t(j)|$ denotes their temporal distance,
and $T$ is the video length.

We then solve the OT problem as follows,
\begin{equation}
\min_{\Gamma\geq 0}
\sum_{i=1}^{m}\sum_{j=1}^{n}
c(i,j)\Gamma(i,j),
\quad
\mathrm{s.t.}\;
\Gamma\mathbf{1}_{n}=\mathbf{b},
\;
\Gamma^{\top}\mathbf{1}_{m}=\mathbf{1}_{n},
\end{equation}
where $\mathbf{b}$ denotes the supplier capacities.
As in~\cite{liu2025boosting}, the capacities are dynamically refined according to the snippet assignments and their updated temporal centers.
Finally, each consumer inherits the label of the supplier receiving the largest transport mass, and the resulting snippet-wise assignments are merged into pseudo temporal segments for training the localization model.

\subsection{Training}

\textbf{Label Assignment.}
The predicted action set is denoted as
$\hat{\Psi}
=
\left\{
\hat{\varphi}_{n}
=
(
\hat{s}_{n}, \hat{e}_{n}, \hat{y}_{n}, \hat{p}_{n}
)
\right\}_{n=1}^{N_{q}}$,
where
$\hat{s}_{n}$ and
$\hat{e}_{n}$
represent the
start and end times
of the $n$-th predicted action instance,
while
$\hat{p}_{n}$ and
$\hat{y}_{n}$
denote the corresponding
class probability distribution and predicted category,
respectively.
The pseudo-label set
$\Psi
=
\left\{
\varphi_i
=
(s_i,e_i,y_i)
\right\}_{i=1}^{N_q}$
includes a padded no-action placeholder, denoted by $\varnothing$.
For a padded no-action label, $y_i=\varnothing$, and its temporal
interval $(s_i,e_i)$ is ignored during temporal matching and regression.
The cost associated with a permutation
$\sigma\in\mathfrak{S}_{N_{q}}$
of the query set is defined as follows,
\begin{align}
\label{Deform}
\mathcal{C}_{cls}^{(i)}(\sigma)
&=
\mathcal{L}_{cls}
\left(
\hat{p}_{\sigma(i)},
y_i
\right),
\\
\mathcal{C}_{iou}^{(i)}(\sigma)
&=
\mathcal{L}_{iou}
\left(
\left(
\hat{s}_{\sigma(i)},
\hat{e}_{\sigma(i)}
\right),
\left(
s_i,
e_i
\right)
\right),
\\
\mathcal{C}_{L_{1}}^{(i)}(\sigma)
&=
\mathcal{L}_{L_{1}}
\left(
\left(
\hat{s}_{\sigma(i)},
\hat{e}_{\sigma(i)}
\right),
\left(
s_i,
e_i
\right)
\right),
\end{align}
where
$y_i$
denotes the class label
of the $i$-th pseudo-label, and $(s_i,e_i)$ denotes an
OT-generated pseudo temporal segment rather than a ground-truth
boundary annotation.
Focal loss~\cite{lin2017focal} is utilized as
$\mathcal{L}_{cls}$
following~\cite{lee2021learning, liu2025stepwise},
while
$\mathcal{L}_{iou}$ and $\mathcal{L}_{L_{1}}$
represent the temporal IoU loss and the $L_{1}$ distance, respectively.
The bipartite matching between two sets aims to find the permutation
$\sigma^{*}$ with the lowest cost,
\begin{equation}
\label{DeformAtt}
\begin{aligned}
\sigma^{*}
=
\underset{\sigma\in\mathfrak{S}_{N_q}}{\arg\min}
\sum_{i=1}^{N_q}
\Big[
&
\alpha_{cls}
\mathcal{C}_{cls}^{(i)}(\sigma)
+
\mathbb{I}(y_i\neq\varnothing)
\Big(\\
&
\alpha_{iou}
\mathcal{C}_{iou}^{(i)}(\sigma)
+
\alpha_{L_{1}}
\mathcal{C}_{L_{1}}^{(i)}(\sigma)
\Big)
\Big],
\end{aligned}
\end{equation}
where $\mathbb{I}(\cdot)$ denotes the indicator function.
The classification cost is computed for all pseudo-labels, including the padded no-action labels, whereas the temporal IoU and $L_1$ costs
are computed only for non-empty pseudo-labels.
The weights
$\alpha_{cls}$, $\alpha_{iou}$ and $\alpha_{L_{1}}$
of the matching costs are
set to 2, 2, and 5, respectively, as in~\cite{kim2024te}.

\textbf{Loss Function.}
After obtaining the best permutation $\sigma^{*}$,
the optimization goal can be expressed as follows,
\begin{equation}
\begin{split}
\mathcal{L}_{\text{detr}} = &\sum_{i=1}^{N_{q}}
\Big[
\lambda_{cls}
\mathcal{C}_{cls}^{(i)}(\sigma^{*}) \\
&+ \mathbb{I}_{\left\{y_{i} \neq \varnothing\right\}}
\big(
\lambda_{iou}
\mathcal{C}_{iou}^{(i)}(\sigma^{*})
+
\lambda_{L_{1}}
\mathcal{C}_{L_{1}}^{(i)}(\sigma^{*})
\big)
\Big],
\end{split}
\end{equation}
where $\mathbb{I}(\cdot)$ is the indicator function and the weights $\lambda_{cls}$, $\lambda_{iou}$ and
$\lambda_{L_{1}}$ are set to 2, 2 and 5, respectively, as in~\cite{kim2024te}.

For the optimization of the pseudo-label generation process, we adopt the focal loss~\cite{lin2017focal} to facilitate the training process.
Let
$\Omega=\Omega_{\mathrm{act}}\cup\Omega_{\mathrm{bg}}$
denote the supervised snippet set, where
$\Omega_{\mathrm{act}}$ contains the annotated action snippets
and $\Omega_{\mathrm{bg}}$ contains the selected background
seeds. For each snippet $t\in\Omega$, let
$y_t^{p}\in\{0,1,\ldots,N_c\}$ denote its pseudo-label, where
$y_t^{p}=0$ represents the background class and
$y_t^{p}\in\{1,\ldots,N_c\}$ represents an action class.
The snippet-level classification loss is defined as follows,
\begin{equation}
\mathcal{L}_{\mathrm{snip}}
=
-\frac{1}{|\Omega|}
\sum_{t\in\Omega}
\left(
1-p_t^{s}(y_t^{p})
\right)^{\beta}
\log p_t^{s}(y_t^{p}),
\label{eq:snippet_loss}
\end{equation}
where $p_t^{s}(y_t^{p})$ denotes the predicted probability that
snippet $t$ belongs to its pseudo-labeled class $y_t^{p}$, and
$\beta$ is the focusing factor, which is set to $2$ following
\cite{lin2017focal}.

% \begingroup
% \color{blue}

 The overall loss function for training our model is as follows,
\begin{equation}
\begin{split}
\label{total loss}
\mathcal{L}_{\text{total}} = \lambda_{detr}\mathcal{L}_{\text{detr}} + \lambda_{act}\mathcal{L}_{\text{snip}} + \lambda_{struct}\mathcal{L}_{\text{struct}},
\end{split}
\end{equation}
where $\lambda_{detr}$, $\lambda_{act}$ and $\lambda_{struct}$ are the weighting parameters to balance the losses, which are set to 1, 0.5 and 1, respectively.
% \endgroup

\section{Experiments and Discussion}
\label{sec:experiments-and-discussions}

In this section, we first introduce four standard benchmarks, the evaluation metrics, and the implementation details. Then, we compare the proposed DeformStruct with the SOTA methods, followed by a set of ablation studies. Note that only point-level labels are leveraged to train the proposed DeformStruct.

\begin{table*}[tb]
\centering
\small
\setlength{\tabcolsep}{3.2pt}
\caption{
Comparison with state-of-the-art methods on THUMOS14 under different supervision settings.
Results of DeformStruct are reported as mean $\pm$ standard deviation over three runs.
}
\label{thumos14}

\begin{tabular}{c|l|ccccc|cc}
\toprule
\multirow{2}{*}{Supervision}
& \multirow{2}{*}{Method}
& \multicolumn{5}{c|}{mAP@tIoU (\%)}
& \multicolumn{2}{c}{AVG} \\
& & 0.3 & 0.4 & 0.5 & 0.6 & 0.7
& [0.1:0.1:0.5] & [0.3:0.1:0.7] \\
\midrule

\multirow{8}{*}{\shortstack{Fully\\supervised}}
& BMN~\cite{lin2019bmn}~(ICCV'19)
& 56.0 & 47.4 & 38.8 & 29.7 & 20.5 & -- & 38.5 \\

& G-TAD~\cite{xu2020g}~(CVPR'20)
& 54.5 & 47.6 & 40.2 & 30.8 & 23.4 & -- & 39.3 \\

& TCANet~\cite{qing2021temporal}~(CVPR'21)
& 60.6 & 53.2 & 44.6 & 36.8 & 26.7 & -- & 44.4 \\

& React~\cite{shi2022react}~(ECCV'22)
& 69.2 & 65.0 & 57.1 & 47.8 & 35.6 & -- & 55.0 \\

& ASL~\cite{shao2023action}~(ICCV'23)
& 83.1 & 79.0 & 71.7 & 59.7 & 45.8 & -- & 67.9 \\

& MCBD~\cite{su2023multi}~(TPAMI'23)
& 68.8 & 62.2 & 53.5 & 41.6 & 29.5 & -- & 51.1 \\

& TE-TAD~\cite{kim2024te}~(CVPR'24)
& 83.3 & 78.4 & 71.3 & 60.7 & 45.6 & -- & 67.9 \\

& AD-IDP~\cite{foo2024action}~(CVPR'24)
& 84.9 & 81.5 & 76.5 & 63.0 & 48.0 & -- & 70.8 \\

\midrule

\multirow{8}{*}{\shortstack{Weakly\\supervised}}
& CMCS~\cite{liu2019completeness}~(CVPR'19)
& 41.2 & 32.1 & 23.1 & 15.0 & 7.0 & 40.9 & 23.7 \\

& DGAM~\cite{shi2020weakly}~(CVPR'20)
& 46.8 & 38.2 & 28.8 & 19.8 & 11.4 & 45.6 & 29.0 \\

& CoLA~\cite{zhang2021cola}~(CVPR'21)
& 51.5 & 41.9 & 32.2 & 22.0 & 13.1 & 50.3 & 32.1 \\

& BackTAL~\cite{yang2021background}~(TPAMI'21)
& 54.4 & 45.5 & 36.3 & 26.2 & 14.8 & -- & 35.4 \\

& STCL-Net~\cite{zhu2023contextloc++}~(TPAMI'23)
& 58.2 & 49.7 & 41.8 & 28.7 & 16.0 & 57.9 & 38.9 \\

& DDG-Net~\cite{tang2023ddg}~(ICCV'23)
& 58.2 & 49.0 & 41.4 & 27.6 & 14.8 & 57.8 & 38.2 \\

& PseudoFormer~\cite{liu2025bridge}~(CVPR'25)
& 65.4 & 55.6 & 44.1 & 31.3 & 17.1 & 62.9 & 42.7 \\

& MLLM4WTAL~\cite{zhang2025weakly}~(CVPR'25)
& 61.8 & 52.3 & 43.0 & 30.8 & 16.6 & 60.2 & 40.9 \\

\midrule

\multirow{10}{*}{\shortstack{Point\\supervised}}
& SF-Net~\cite{ma2020sf}~(ECCV'20)
& 53.2 & 40.7 & 29.3 & 18.4 & 9.6 & 51.5 & 30.2 \\

& Ju \textit{et al.}~\cite{ju2021divide}~(ICCV'21)
& 58.1 & 46.4 & 34.5 & 21.8 & 11.9 & 55.3 & 34.5 \\

& LACP~\cite{lee2021learning}~(ICCV'21)
& 63.3 & 55.2 & 43.9 & 33.3 & 20.8 & 61.6 & 43.3 \\

& HR-Pro~\cite{zhang2024hr}~(AAAI'24)
& 74.3 & 64.3 & 52.2 & 39.8 & 24.8 & 71.6 & 51.1 \\

& RCTSI~\cite{xia2024realigning}~(CVPR'24)
& 70.1 & 60.0 & 49.4 & 37.6 & 24.5 & 67.9 & 48.3 \\

& SMBD~\cite{liu2025stepwise}~(ECCV'24)
& 66.0 & 57.9 & 47.0 & 36.0 & 22.0 & 64.2 & 45.7 \\

& CRC~\cite{yang2025cross}~(TPAMI'25)
& 65.5 & 56.9 & 45.4 & 34.9 & 21.2 & 63.2 & 44.8 \\

\rowcolor{green!10}
& QROT~\cite{liu2025boosting}~(CVPR'25)
& 73.1 & 64.4 & 54.3 & 41.3 & 27.4 & 72.3 & 52.1 \\

\rowcolor{cyan!10}
& DeformStruct w/o RSO
& ${75.6}^{\scriptscriptstyle\pm0.7}$
& ${67.8}^{\scriptscriptstyle\pm0.8}$
& ${57.4}^{\scriptscriptstyle\pm0.5}$
& ${42.8}^{\scriptscriptstyle\pm0.4}$
& $27.8^{\scriptscriptstyle\pm0.7}$
& ${73.5}^{\scriptscriptstyle\pm0.6}$
& ${54.3}^{\scriptscriptstyle\pm0.5}$ \\
\rowcolor{cyan!10}
& DeformStruct (Ours)
& $\mathbf{76.4}^{\scriptscriptstyle\pm0.5}$
& $\mathbf{68.6}^{\scriptscriptstyle\pm0.6}$
& $\mathbf{57.9}^{\scriptscriptstyle\pm0.8}$
& $\mathbf{43.2}^{\scriptscriptstyle\pm0.9}$
& $\mathbf{28.0}^{\scriptscriptstyle\pm0.4}$
& $\mathbf{74.1}^{\scriptscriptstyle\pm0.7}$
& $\mathbf{54.8}^{\scriptscriptstyle\pm0.8}$ \\

\bottomrule
\end{tabular}
\end{table*}

\subsection{Dataset and Evaluation}
\noindent\textbf{THUMOS14.} The training data in THUMOS14~\cite{jiang2014thumos} is based on the UCF101~\cite{soomro2012ucf101} action dataset. There are a total of 200 validation videos and 213 test videos that belong to 20 action classes. On average, each video in the dataset includes 15 action instances, making the task of recognizing all actions quite challenging. Similarly to~\cite{ma2020sf}, we use the validation set for training and the test set for evaluation.\\
\noindent\textbf{ActivityNet-v1.3.} 
The ActivityNet-v1.3~\cite{caba2015activitynet} dataset includes 200 different types of activities, totaling 849 hours of YouTube video. In terms of activity categories and video quantity, ActivityNet-v1.3 is currently the largest benchmark for temporal activity detection. ActivityNet-v1.3 contains a total of 19,994 untrimmed videos, divided into training, validation, and test subsets in a 2:1:1 ratio. Each activity category includes an average of 137 untrimmed videos, with each video containing an average of 1.41 activities.\\
\noindent\textbf{GTEA.} The GTEA~\cite{lei2018temporal} comprises 28 videos and contains seven different types of everyday activities. Each video lasts approximately one minute and contains around 20 annotated action instances. For our experiments, we used 21 videos for training and 7 for testing. \\
\noindent
\textbf{BEOID.} The BEOID, as described in~\cite{damen2014you}, comprises 58 videos representing 30 action classes, and each video in the dataset contains an average of 12.5 actions. Following~\cite{ma2020sf}, $80\%$ of the videos are used for training, while the remaining $20\%$ are reserved for testing.\\

\noindent\textbf{Evaluation Metrics.}
Following previous PTAL works~\cite{ma2020sf,ju2021divide,lee2021learning, zhang2024hr, xia2024realigning, liu2025stepwise}, we adopt the standard evaluation protocol for all datasets.
For THUMOS14, we report mAP at IoU thresholds from 0.3 to 0.7
and the average mAP over both $[0.1:0.1:0.5]$ and
$[0.3:0.1:0.7]$. For GTEA and BEOID, the average mAP is
calculated over $[0.1:0.1:0.7]$. For ActivityNet-v1.3, we report
mAP at 0.5, 0.75, and 0.95, together with the average over
$[0.5:0.05:0.95]$.

\begin{figure*}[!t]
\centering
	\includegraphics[width=\textwidth]{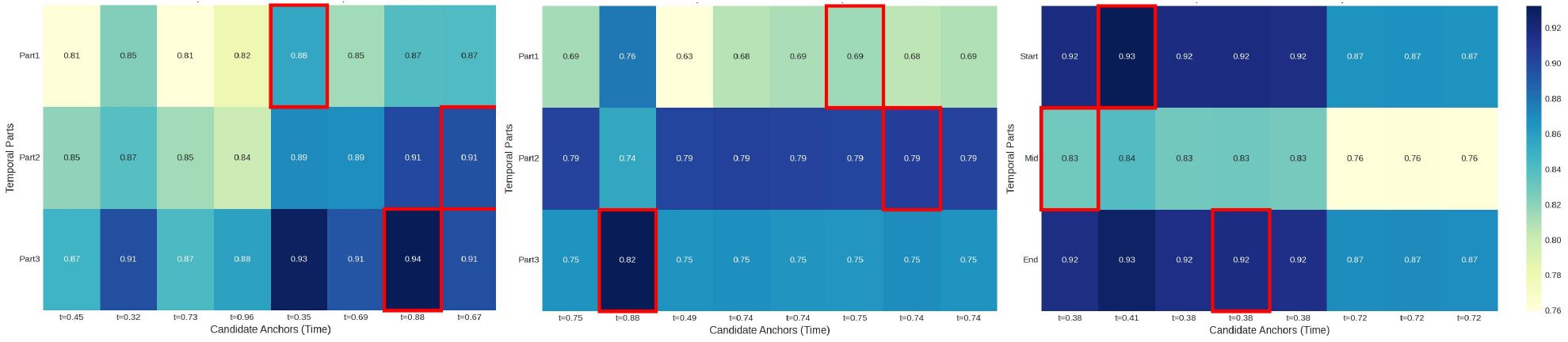}
	\caption{Visualization of the Part-Anchor alignment score $V$. The rows represent the learned deformable parts, and the columns correspond to the candidate anchors mined from the video. The color intensity indicates the matching score, where darker colors represent higher compatibility. The red boxes highlight the optimal assignment results obtained by the Hungarian algorithm. This matrix demonstrates how our model jointly considers appearance similarity and temporal deformation to align parts with anchors.}
  \label{costmatrix}
\end{figure*}

\subsection{Implementation Details}

We use a two-stream I3D network~\cite{carreira2017quo} pretrained on Kinetics-400~\cite{carreira2017quo} as the feature extractor. We divide each video into non-overlapping 16-frame snippets and apply the TV-L1 algorithm~\cite{wedel2009improved} to extract optical flow.
After obtaining the RGB and optical-flow features, we employ the two-stream fusion operation as described in~\cite{narayan20193c} to integrate the RGB and optical-flow branches, resulting in a 2048-dimensional vector for each snippet. The number of non-overlapping snippets is denoted as $T$. For all datasets, we optimize our method using Adam~\cite{kingma2014adam} with a learning rate of $10^{-4}$ and a batch size of 8. The numbers of encoder and decoder layers, \(L_E\) and \(L_D\), are set to 4 and 6, respectively. The number of attention heads \(M\) is set to 8, and the number of sampling points \(K_s\) is set to 4.
All results for our framework use the same settings as QROT~\cite{liu2025boosting}, including I3D features, point-label protocol, data splits, and evaluation code. We adopt TE-TAD~\cite{kim2024te} as the base framework and adapt it to point supervision as our base model. The SOTA comparison on THUMOS14 is reported as mean $\pm$ standard deviation over three runs, whereas the ablation studies report the mean performance over three runs.
\begin{figure*}[!t]
\centering
	\includegraphics[width=\textwidth]{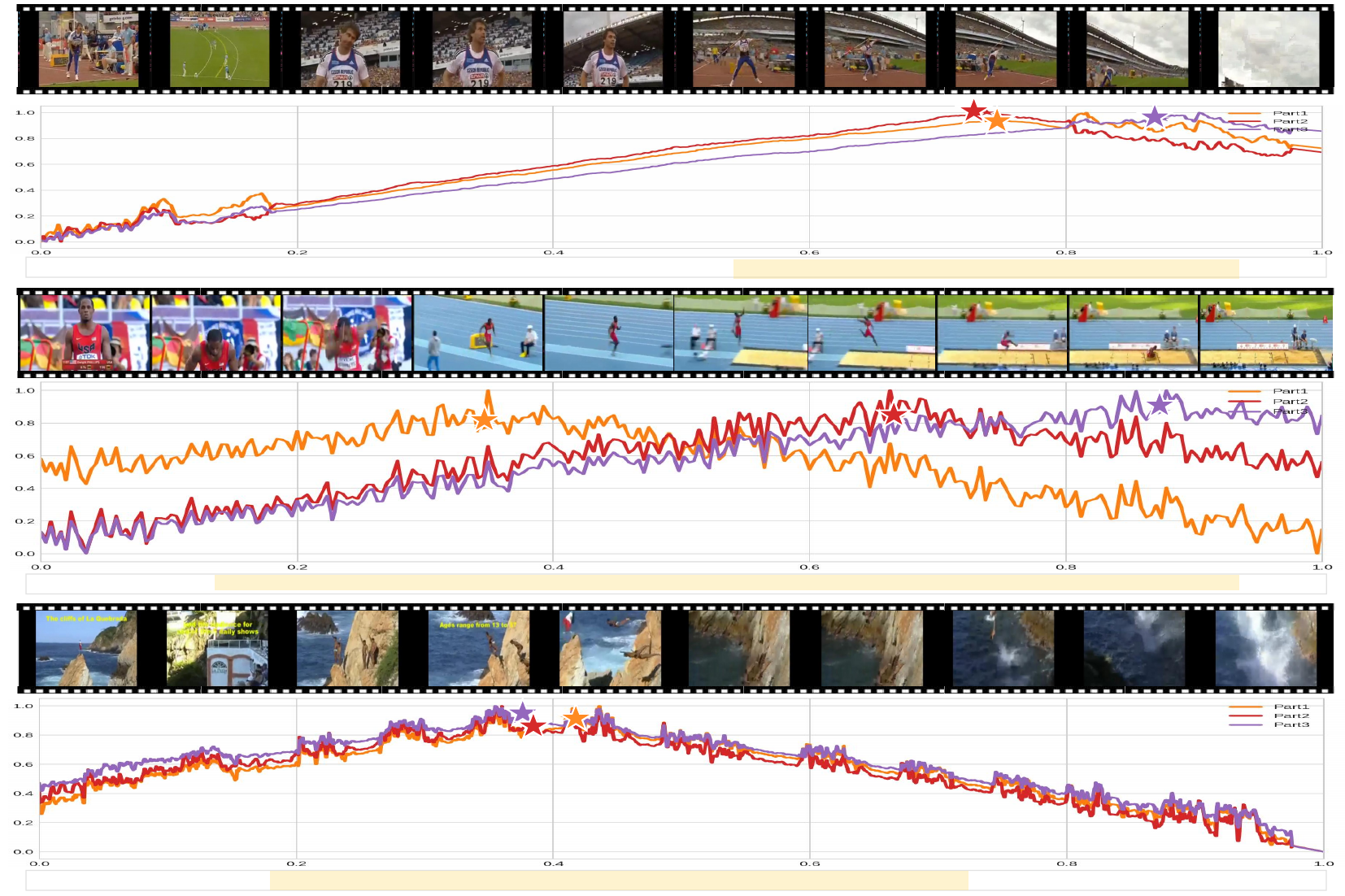}
	\caption{Qualitative visualization of the DPA module on THUMOS14. The colored markers indicate the locations of anchors matched with latent action parts. The bars at the bottom indicate the ground-truth segments. Notably, the mined anchors align well with the ground-truth action intervals and capture complementary action-relevant regions distributed over time.}
  \label{part_visual}
\end{figure*}
\begin{table}[t]
\centering
\small
\setlength{\tabcolsep}{4.5pt}
\caption{
Comparison with the state-of-the-art point-supervised methods on ActivityNet-v1.3. AVG represents the mean average precision (mAP) at thresholds $[0.5:0.05:0.95]$.
}
\label{ActivityNet}

\begin{tabular}{l|ccc|c}
\toprule
\multirow{2}{*}{Method}
& \multicolumn{3}{c|}{mAP@tIoU (\%)}
& \multicolumn{1}{c}{AVG} \\
& 0.5 & 0.75 & 0.95 & [0.5:0.05:0.95] \\
\midrule

SF-Net~\cite{ma2020sf}
& -- & -- & -- & -- \\

Ju~\textit{et al.}~\cite{ju2021divide}
& -- & -- & -- & -- \\

LACP~\cite{lee2021learning}
& 40.4 & 24.6 & 5.7 & 25.1 \\

HR-Pro~\cite{zhang2024hr}
& 42.8 & 27.2 & 8.0 & 27.1 \\

RCTSI~\cite{xia2024realigning}
& -- & -- & -- & -- \\

SMBD~\cite{liu2025stepwise}
& 43.2 & 27.8 & 8.3 & 27.5 \\

\rowcolor{green!10}
QROT~\cite{liu2025boosting}
& 43.3 & 28.2 & 8.5 & 27.7 \\
\rowcolor{cyan!10}
DeformStruct w/o RSO
& 43.8 & 28.6 & 8.9 & 28.1 \\

\rowcolor{cyan!10}
DeformStruct (Ours)
& \textbf{44.1}
& \textbf{29.2}
& \textbf{9.3}
& \textbf{28.5} \\

\bottomrule
\end{tabular}
\end{table}

\begin{table}[t]
\centering
\small
\setlength{\tabcolsep}{4.5pt}
\caption{
Comparison with state-of-the-art point-supervised methods on GTEA.
AVG represents the mean average precision (mAP) at thresholds $[0.1:0.1:0.7]$.
}
\label{GTEA}
\begin{tabular}{l|ccc|c}
\toprule
\multirow{2}{*}{Method}
& \multicolumn{3}{c|}{mAP@tIoU (\%)}
 & \multicolumn{1}{c}{AVG} \\ 
 &0.3&0.5&0.7 &  [0.1:0.1:0.7] \\ 
\midrule

SF-Net~\cite{ma2020sf}
& 37.9 & 19.3 & 11.9 & 31.0 \\

Ju \textit{et al.}~\cite{ju2021divide}
& 38.3 & 21.9 & 18.1 & 33.7 \\

LACP~\cite{lee2021learning}
& 55.7 & 33.9 & 20.8 & 43.5 \\

HR-Pro~\cite{zhang2024hr}
& 61.1 & 37.3 & 17.5 & 47.3 \\

RCTSI~\cite{xia2024realigning}
& 60.9 & 39.5 & 16.6 & 49.0 \\

SMBD~\cite{liu2025stepwise}
& 61.3 & 41.1 & 14.2 & 47.4 \\

\rowcolor{green!10}
QROT~\cite{liu2025boosting}
& 61.5 & 37.9 & 18.2 & 47.9 \\
\rowcolor{cyan!10}
DeformStruct w/o RSO
& 64.2 & 40.1 & 18.5 & 49.1 \\

\rowcolor{cyan!10}
DeformStruct (Ours)
& \textbf{64.7}
& \textbf{40.9}
& \textbf{19.3}
& \textbf{49.8} \\

\bottomrule
\end{tabular}
\end{table}

% \begin{table}[t]
%   \caption{\textcolor{blue}{Comparison with the state-of-the-art methods on GTEA. AVG represents the mean average precision (mAP) at thresholds $[0.3:0.1:0.7]$.}}
%   \label{GTEA}
%   \centering
%   \small
% \begin{tabular}{c|l|ccc|c}
% \toprule
% \multirow{2}{*}{Dataset} & \multirow{2}{*}{Method} & \multicolumn{3}{c|}{mAP@tIoU (\%)} & \multicolumn{1}{c}{AVG} \\ 
%  & &0.3&0.5&0.7 &  [0.3:0.1:0.7] \\ 
% \midrule
% \multirow{8}{*}{GTEA} & SF-Net~\cite{ma2020sf}   & 37.9  & 19.3  & 11.9 & 31.0 \\
% & Ju \textit{et al.}~\cite{ju2021divide}  & 38.3 & 21.9 & 18.1 & 33.7  \\
% & LACP~\cite{lee2021learning}  & 55.7 & 33.9 & 20.8 & 43.5 \\
% & HR-Pro~\cite{zhang2024hr}  & 61.1 & 37.3& 17.5& 47.3\\
% & RCTSI~\cite{xia2024realigning}   & 60.9 & 39.5 & 16.6 & 49.0  \\ 
% & SMBD~\cite{liu2025stepwise}&  61.3 & 41.1 & 14.2 & 47.4\\
% &\cellcolor{green!10}QROT~\cite{liu2025boosting} &\cellcolor{green!10}61.5 & \cellcolor{green!10}37.9  & \cellcolor{green!10}18.2 & \cellcolor{green!10}47.9 \\
% &\cellcolor{cyan!10}QROT+DPA  & \cellcolor{cyan!10}64.2 & \cellcolor{cyan!10}40.1&  \cellcolor{cyan!10}18.5& \cellcolor{cyan!10}49.1\\
% &\cellcolor{cyan!10}QROT+DPA+RSO  & \cellcolor{cyan!10}64.7 & \cellcolor{cyan!10}40.9&  \cellcolor{cyan!10}19.3& \cellcolor{cyan!10}49.8\\
% \bottomrule
%   \end{tabular}
% \end{table}

\begin{table}[t]
\centering
\small
\setlength{\tabcolsep}{4.5pt}
\caption{
Comparison with the state-of-the-art point-supervised methods on BEOID. AVG represents the mean average precision (mAP) at thresholds $[0.1:0.1:0.7]$.
}
\label{BEOID}

\begin{tabular}{l|ccc|c}
\toprule
\multirow{2}{*}{Method}
& \multicolumn{3}{c|}{mAP@tIoU (\%)}
& \multicolumn{1}{c}{AVG} \\
& 0.3 & 0.5 & 0.7 & [0.1:0.1:0.7] \\
\midrule

SF-Net~\cite{ma2020sf}
& 40.6 & 16.7 & 3.5 & 30.9 \\

Ju \textit{et al.}~\cite{ju2021divide}
& 46.8 & 20.9 & 5.8 & 34.9 \\

LACP~\cite{lee2021learning}
& 61.4 & 42.7 & 25.1 & 51.8 \\

HR-Pro~\cite{zhang2024hr}
& 72.1 & 55.3 & 26.1 & 59.4 \\

RCTSI~\cite{xia2024realigning}
& 73.0 & 51.1 & 23.8 & 59.6 \\

SMBD~\cite{liu2025stepwise}
& 71.0 & 52.5 & 25.2 & 57.4 \\

\rowcolor{green!10}
QROT~\cite{liu2025boosting}
& 73.0 & 55.9 & 26.7 & 59.9 \\
\rowcolor{cyan!10}
DeformStruct w/o RSO
& 74.9 & 56.7 & 27.3 & 60.7 \\

\rowcolor{cyan!10}
DeformStruct (Ours)
& \textbf{75.7}
& \textbf{57.8}
& \textbf{28.1}
& \textbf{61.4} \\

\bottomrule
\end{tabular}
\end{table}

% \begingroup
% \color{blue}
\subsection{Comparisons with the state-of-the-art methods}
\noindent\textbf{THUMOS14.}
Table~\ref{thumos14} compares DeformStruct with fully-supervised, weakly-supervised, and point-supervised methods on THUMOS14.
Under point supervision, our conference version QROT already achieves competitive localization performance.
Building upon it, introducing deformable part modeling consistently improves the overall performance, demonstrating the effectiveness of capturing multiple action-related parts.
Further incorporating RSO brings additional gains, indicating that modeling the relative organization among aligned parts provides complementary structural information beyond independent part alignment.
Overall, DeformStruct achieves the best average performance among the compared point-supervised methods, while the results over three runs further characterize the stability of the model across different runs.

\noindent\textbf{ActivityNet-v1.3.}
As shown in Table~\ref{ActivityNet}, DeformStruct consistently outperforms previous point-supervised methods, including QROT, across all reported tIoU thresholds. Further incorporating RSO leads to additional improvements, indicating that relative temporal organization provides complementary cues for action localization.

\noindent\textbf{GTEA.}
On GTEA, DeformStruct achieves the best average performance among the compared point-supervised methods, as reported in Table~\ref{GTEA}. The gain over the variant without RSO further supports the usefulness of relational structure modeling for capturing the organization of aligned action parts.

\noindent\textbf{BEOID.}
Table~\ref{BEOID} shows that DeformStruct improves upon existing PTAL approaches across the reported tIoU thresholds. Moreover, the full model consistently surpasses DeformStruct w/o RSO, suggesting that relational modeling contributes structural information beyond deformable part alignment.

Overall, DeformStruct consistently outperforms state-of-the-art point-supervised methods across all four datasets, including our conference version QROT~\cite{liu2025boosting}.
These results demonstrate the general effectiveness of the proposed deformable structural modeling across diverse action datasets.

\begin{figure*}[ht]
\centering
\includegraphics[width=\linewidth]{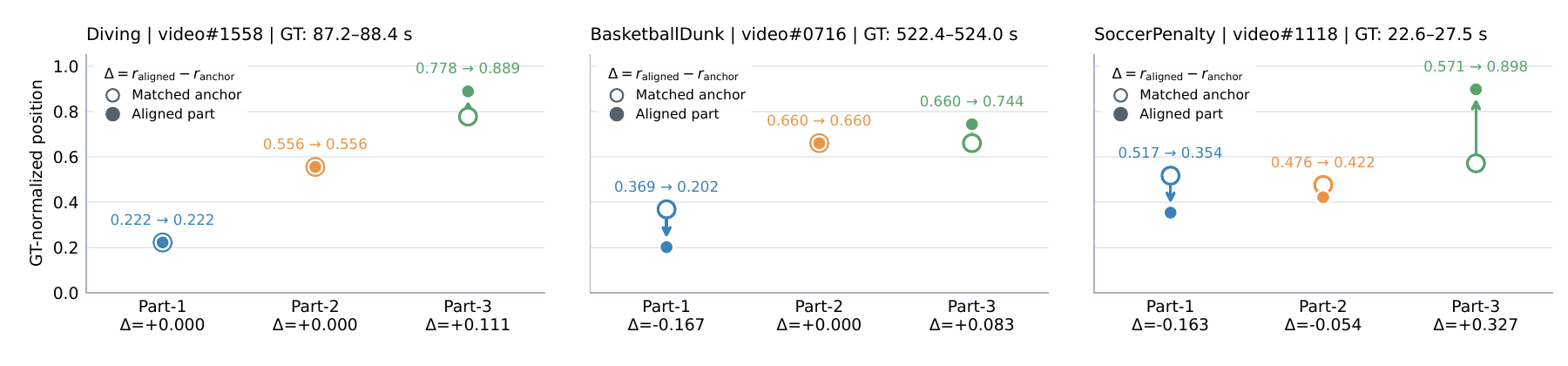}
\caption{
Visualization of instance-adaptive part deformation.
For each action instance, hollow and solid circles denote the matched anchor positions and the final aligned part positions, respectively, both normalized within the ground-truth action interval.
Arrows indicate the deformation from anchors to aligned parts, where $\Delta = r_{\mathrm{aligned}} - r_{\mathrm{anchor}}$.
Different latent parts exhibit instance-dependent shifts in both direction and magnitude.
}
\label{fig:instance_deformations}
\end{figure*}

% }

\subsection{Ablation Study}
\label{subsec:ablation-study}

In this subsection, to better analyze the contribution of each component, we conduct ablation studies on THUMOS14. For clarity, we highlight the conference version in green and the final setting used for comparison with SOTA methods in blue.

\noindent\textbf{Ablation study of different modules on THUMOS14.}
Table~\ref{tab:strong_detr_ablation_thumos1} first compares different pseudo-label generation strategies under the same DETR architecture. Specifically, NPE denotes no pseudo-label expansion, FW and GW represent fixed-window and Gaussian-window expansion, respectively. SF-Net follows the adaptive pseudo-label mining strategy of SF-Net~\cite{ma2020sf}, GS denotes greedy search in LACP~\cite{lee2021learning}, and OT denotes optimal transport. Among these strategies, OT achieves the best overall performance.
Building upon OT, both QR and DPA improve localization performance, and their combination further raises the average mAP to 54.3\%, showing their complementary benefits.
Adding RSO further improves the average mAP to 54.8\% and yields gains across all reported tIoU thresholds.
These results verify that modeling the relative temporal organization among aligned parts provides additional structural cues beyond deformable part alignment.

\begin{table}[t]
\centering
\small
\setlength{\tabcolsep}{4.5pt}
\caption{Comparison of different pseudo-label generation strategies and component contributions under the same DETR architecture on THUMOS14.
}
\label{tab:strong_detr_ablation_thumos1}
\begin{tabular}{l|ccc|ccc|c}
\toprule
\multirow{2}{*}{Setting}
& \multirow{2}{*}{QR}
& \multirow{2}{*}{DPA}
& \multirow{2}{*}{RSO}
& \multicolumn{3}{c|}{mAP@tIoU (\%)}
& \multicolumn{1}{c}{AVG} \\
& & & & 0.3 & 0.5 & 0.7 & [0.3:0.1:0.7] \\
\midrule

\multicolumn{8}{c}{\textit{Pseudo-label Generation}} \\
\midrule
NPE    & -- & -- & -- & 4.5  & 1.9  & 0.4  & 2.2  \\
FW     & -- & -- & -- & 35.9 & 13.9 & 2.4  & 16.4 \\
GW     & -- & -- & -- & 12.2 & 5.4  & 1.3  & 5.9  \\
% CE     & -- & -- & -- & 56.8 & 29.2 & 7.2  & 30.4 \\
SF-Net & -- & -- & -- & 54.3 & 31.4 & 10.7 & 31.6 \\
GS     & -- & -- & -- & 71.9 & 50.6 & 25.0 & 49.4 \\
OT     & -- & -- & -- & 72.0 & 53.4 & 25.2 & 50.9 \\

\midrule
\multicolumn{8}{c}{\textit{Component Ablation with OT Supervision}} \\
\midrule
OT
& -- & -- & --
& 72.0 & 53.4 & 25.2 & 50.9 \\

\rowcolor{green!10}OT
& \checkmark & -- & --
& 73.1 & 54.3 & 27.4 & 52.1 \\

OT
& -- & \checkmark & --
& 74.9 & 56.7 & 22.8 & 53.0 \\

OT
& \checkmark & \checkmark & --
& 75.6 & 57.4 & 27.8 & 54.3 \\

\rowcolor{cyan!10}
OT
& \checkmark & \checkmark & \checkmark
& \textbf{76.4} & \textbf{57.9} & \textbf{28.0} & \textbf{54.8 }\\

\bottomrule
\end{tabular}
\end{table}

% \begin{table}[h]
% \centering
% \small
% \caption{\textcolor{blue}{
% Comparison of different pseudo-label generation strategies under the non-DETR architecture. 
% }}
% \label{tab:pseudo_label_generation_comparison1}
% \begin{tabular}{l|ccc|c}
% \toprule
%  \multirow{2}{*}{Method} & \multicolumn{3}{c|}{mAP@tIoU (\%)} & \multicolumn{1}{c}{AVG} \\ 
%  &0.3&0.5&0.7 &  [0.3:0.1:0.7] \\ 
% \midrule
% SF-Net & 52.8 & 30.5 & 12.0 & 31.6 \\
% GA~(LACP)  & 63.3 & 43.9 &  20.8  & 43.3 \\
% % & 64.6 & 45.3 & 21.8 & 44.5 \\
% OT & 65.7 & 47.1 & 22.4 & 45.4 \\
% \bottomrule
% \end{tabular}
% \end{table}

\begin{table}[t]
	\centering
    \small
\caption{
Sensitivity analysis of the loss weights on THUMOS14. We evaluate the effects of $\lambda_{\text{part}}$, $\lambda_{\text{align}}$, and $\lambda_{\text{rel}}$ on localization performance.
}
	\label{L_DPA}
	% \resizebox*{\linewidth}{!}{
		\begin{tabular}{l|c|ccc|c}
         \toprule
			\multirow{2}{*}{Loss} & \multirow{2}{*}{Weight}  &
			\multicolumn{3}{c|}{mAP@tIoU (\%)} & \multicolumn{1}{c}{AVG}   \\
			 & & 0.3 & 0.5 & 0.7 & [0.3:0.1:0.7] \\
		\midrule
			
			% $\pi^{*}$  &  &  &  &  &   \\
			% \hline
			\multirow{4}{*}{$\lambda_{\text{part}}$} 
            & 0.1 & 73.7 & 51.9 & 20.4 & 49.2 \\
			& 0.5 & 75.2 & 56.1 & 25.5 & 53.4  \\
			&  1\cellcolor{cyan!10} & \textbf{76.4}\cellcolor{cyan!10} & \textbf{57.9}\cellcolor{cyan!10} & \textbf{28.0}\cellcolor{cyan!10} & \textbf{54.8}\cellcolor{cyan!10} \\
			&  2 & 75.5 & 56.0 & 24.0 & 53.3  \\
		\midrule
			\multirow{4}{*}{$\lambda_{\text{align}}$} 
            & 0.1  & 75.1 & 57.3 & 26.0 & 53.9  \\
			&  0.5\cellcolor{cyan!10} & \textbf{76.4}\cellcolor{cyan!10} & \textbf{57.9}\cellcolor{cyan!10} & \textbf{28.0}\cellcolor{cyan!10} & \textbf{54.8}\cellcolor{cyan!10} \\
			 & 1 & 75.0 & 56.7& 22.3& 52.7  \\
		      & 2  & 75.5 & 53.8 & 21.5 & 51.4  \\
        \midrule
        \multirow{4}{*}{$\lambda_{\text{rel}}$} 
            & 0.1  & 76.7 & 55.6 & 23.8& 53.1 \\
			&  0.5 & 76.5 & 57.0 & 25.6& 54.0 \\
			& 1\cellcolor{cyan!10} & \textbf{76.4}\cellcolor{cyan!10} & \textbf{57.9}\cellcolor{cyan!10} & \textbf{28.0}\cellcolor{cyan!10} & \textbf{54.8}\cellcolor{cyan!10} \\
		    & 2 & 76.4 & 55.6 & 25.5& 53.3 \\
			\bottomrule
		\end{tabular}
	% }
\end{table}

% \begin{table}[t]
% 	\centering
%     \small
% 	\caption{Ablation study of the loss terms in the QROT module on the THUMOS14 dataset}
% 	\label{L_act_detr}
% 	% \resizebox*{\linewidth}{!}{
% 		\begin{tabular}{l|c|ccc|c}
%           \toprule
% 			\multirow{2}{*}{QROT loss} & \multirow{2}{*}{$Num.$}  &
% 			\multicolumn{3}{c|}{mAP@tIoU (\%)} & \multicolumn{1}{c}{AVG}   \\
% 			 & & 0.3 & 0.5 & 0.7 & [0.3:0.1:0.7] \\
% 			\midrule
% 			  \multirow{4}{*}{$\lambda_{act}$}& 0.05 & 71.9  & 53.0 & 23.5 & 52.3 \\
% 		     & 0.1 &  71.2   & 51.1   &  23.7 & 49.3 \\
% 			& \cellcolor{cyan!10}0.5  & \cellcolor{cyan!10}73.1  &\cellcolor{cyan!10}54.3 &\cellcolor{cyan!10}27.4 &\cellcolor{cyan!10}52.1  \\
% 			  & 1  &  72.0  & 50.5 & 24.9 & 49.5 \\
			
% 		\bottomrule
% 		\end{tabular}
% 	% }
% \end{table}

\noindent
\textbf{Visualization of the Part-Anchor Alignment Score Matrix.} In Fig.~\ref{costmatrix}, we provide a visualization of the Part-Anchor Alignment Scores in the DeformStruct module. The rows represent the learned deformable parts (Part 1 to Part 3), and the columns correspond to the candidate anchors mined from the video. The color intensity indicates the matching score (negative cost), where darker colors represent higher compatibility. The red boxes highlight the optimal assignment results obtained by the Hungarian algorithm. This matrix demonstrates how our model jointly considers appearance similarity and temporal deformation to align parts with anchors.

\noindent
\textbf{Ablation study of the loss weights in DeformStruct.} In Table~\ref{L_DPA}, we present an ablation study of the loss weights. Rows 1 to 4 correspond to different settings of $\lambda_{\text{part}}$, rows 5 to 8 report the results of varying $\lambda_{\text{align}}$, and the final four rows evaluate different settings of $\lambda_{\text{rel}}$ for RSO.

\noindent\textbf{Effect of Structural Consistency Objectives.}
We evaluate the contribution of each objective in Table~\ref{tab:structural_loss_ablation}. The performance consistently decreases when any individual objective is removed, demonstrating that the three losses provide complementary supervision for learning the action structure.
% In particular, the relational consistency loss brings more evident gains at stricter tIoU thresholds, highlighting the importance of preserving the relative temporal organization of aligned parts. The full objective achieves the best overall performance.

\begin{table}[ht]
    \centering
    \small
    \caption{Ablation study of different structural consistency objectives on THUMOS14.}
    \label{tab:structural_loss_ablation}
    \begin{tabular}{ccc|ccc|c}
        \toprule
        \multirow{2}{*}{$\mathcal{L}_{\mathrm{part}}$} &
        \multirow{2}{*}{$\mathcal{L}_{\mathrm{align}}$} &
        \multirow{2}{*}{$\mathcal{L}_{\mathrm{rel}}$} &
        \multicolumn{3}{c|}{mAP@tIoU (\%)} &
        \multicolumn{1}{c}{AVG} \\
        & & & 0.3 & 0.5 & 0.7 & [0.3:0.1:0.7] \\
        \midrule
        \checkmark & \checkmark &  
        & 75.6 & 57.4 & 27.8 & 54.3 \\

         & \checkmark & \checkmark &
        75.0 & 57.1 & 26.8 & 53.1 \\

        \checkmark &  & \checkmark &
        75.1 & 56.9 & 27.1 & 53.5 \\

        \rowcolor{cyan!10}
        \checkmark & \checkmark & \checkmark
       & \textbf{76.4} & \textbf{57.9} & \textbf{28.0}& \textbf{54.8} \\
        \bottomrule
    \end{tabular}
\end{table}

\begin{figure*}[t]
    \centering
    \includegraphics[width=0.95\linewidth]{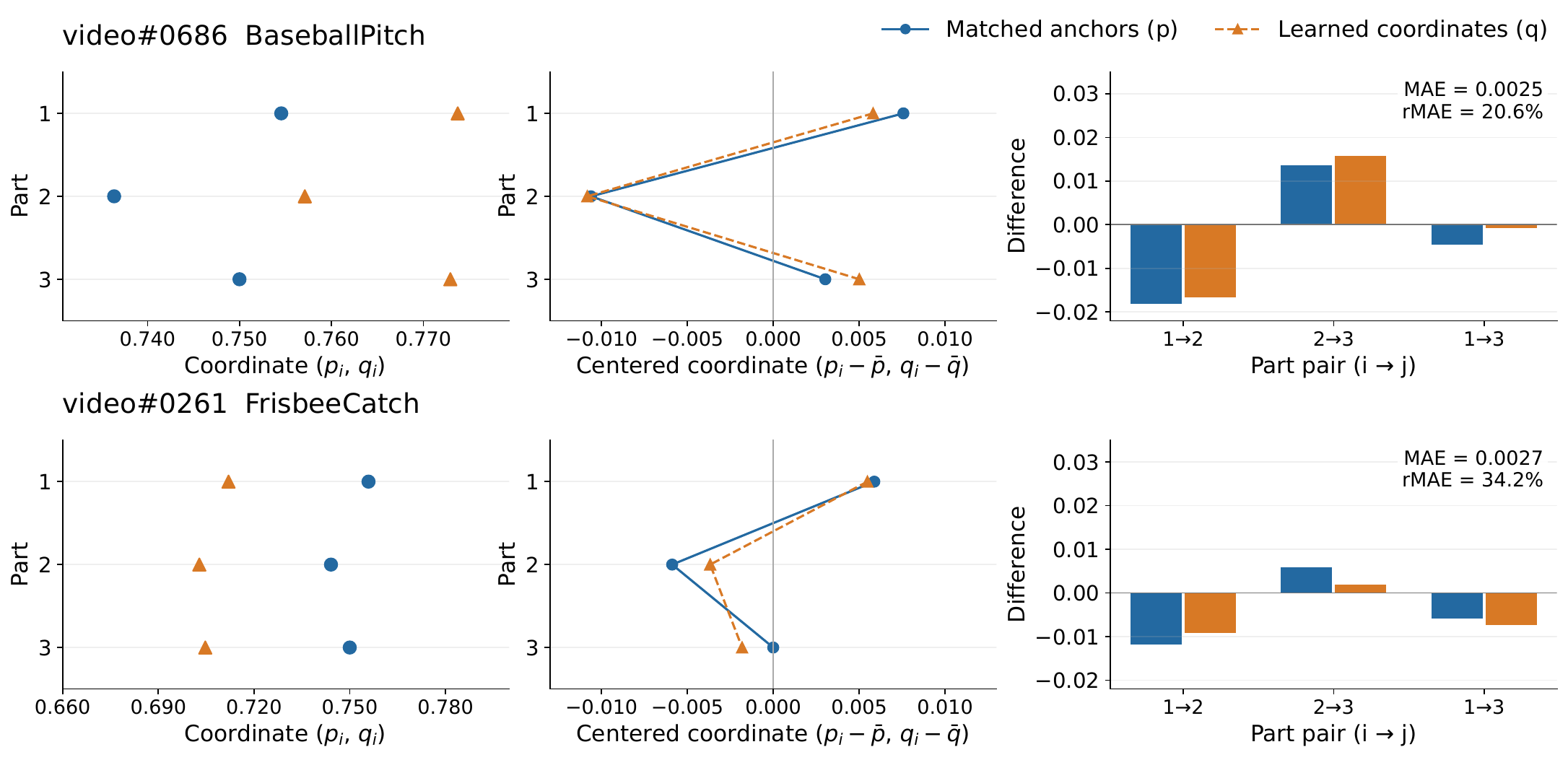}
    \caption{Qualitative visualization of relational structures learned by RSO. From left to right: aligned-part coordinates and learned structural coordinates, their mean-centered configurations, and pairwise coordinate differences. These examples show that RSO preserves part ordering and approximately captures relative spacing despite absolute coordinate offsets, supporting its
  ability to learn relational structure.}
   \label{fig: rso}
\end{figure*}

\noindent\textbf{Effect of Relational Structure Modeling.}
To examine whether the improvement of RSO is merely due to additional
temporal-position supervision, we compare several structural formulations while keeping the DPA framework unchanged. As shown in Table~\ref{tab:rso_relation_ablation}, directly regressing the absolute positions of aligned parts does not improve the overall performance over DPA, indicating that absolute temporal-position supervision alone is insufficient to characterize the structural organization of action parts. Modeling only pairwise temporal distances also leads to degraded performance, suggesting that separation without ordering cannot adequately describe their temporal organization. Moreover, randomized relational targets clearly degrade localization performance, showing that incorrect structural relations can be harmful. In contrast, RSO achieves the best overall performance by jointly preserving the temporal ordering and relative separation among aligned parts.

\begin{table}[ht]
    \centering
    \small
    \caption{Ablation study of different relational formulations.}
    \label{tab:rso_relation_ablation}
    \begin{tabular}{l|ccc|c}
        \toprule
       \multirow{2}{*}{Variant}
        & \multicolumn{3}{c|}{mAP@tIoU (\%)} 
        & \multicolumn{1}{c}{AVG} \\
        & 0.3 & 0.5 & 0.7 & [0.3:0.1:0.7] \\
        \midrule
        DPA
        & 75.6 & 57.4 & 27.8 & 54.3 \\
        
        + Absolute Position
        & 75.8 & 57.7 & 26.6 & 53.8 \\

        + Distance Relation
        & 76.1 & 55.2 & 22.6 & 52.2 \\

        + Randomized Relation
        & 74.1 & 54.1 & 21.5 & 51.1 \\
\rowcolor{cyan!10}
        +  RSO
          & \textbf{76.4} &\textbf{57.9} & \textbf{28.0}& \textbf{54.8} \\
        \bottomrule
    \end{tabular}
\end{table}
% \begin{table}[ht]
%     \centering
%     \small
%     \caption{Ablation study of different relational structure modeling strategies on THUMOS14.}
%     \label{tab:rso_relation_ablation}
%     \begin{tabular}{l|l|ccc|c}
%         \toprule
%         Method & Relation & \multicolumn{3}{c|}{mAP@tIoU (\%)} & \multicolumn{1}{c}{AVG}  \\
%         & & 0.3  & 0.5  & 0.7 & [0.3:0.1:0.7] \\
%         \midrule
%         DPA & None
%         & 75.4 & 57.5 & 24.3 & 54.3 \\
%         DPA + RSO & Absolute
%         & 76.1  & 55.2  & 22.6 & 52.2\\ 
%         DPA + RSO & Shuffled
%         & 74.1  & 54.1 & 21.5 & 51.1\\  
%         DPA + RSO & Signed
%         & 76.2 & 58.2 & 26.3 & 54.8   \\
%         \bottomrule
%     \end{tabular}
% \end{table}

\noindent\textbf{Effect of Relational Representation.}
We further compare different strategies for representing the relations among aligned parts. MAE measures the error in predicting pairwise temporal differences, Order Acc. evaluates whether the relative temporal order between parts is correctly preserved, and Spearman $\rho$ measures the consistency of the overall relational ranking. As shown in Table~\ref{tab:rso_relation_formulation}, Shared Coord. achieves a slightly lower MAE, while substantially improving Order Acc. from 55.7\% to 69.5\% and Spearman $\rho$ from 0.201 to 0.397, together with a higher average mAP. These results indicate that the shared-coordinate formulation is more effective at preserving the global temporal organization among action parts, rather than independently fitting individual pairwise relations.

\begin{table}[ht]
\centering
\small
\setlength{\tabcolsep}{2.8pt}
\caption{Comparison of relational representation strategies on THUMOS14. MAE measures pairwise relation error, while Order Acc. and Spearman $\rho$ evaluate relative temporal structure.}
\label{tab:rso_relation_formulation}
\begin{tabular}{lccccc}
\toprule
Relation & Param. & MAE $\downarrow$ & Order Acc. $\uparrow$ & $\rho$ $\uparrow$ & AVG \\
\midrule
Pairwise Rel.
& 197,121
& 0.1294
& 55.7
& 0.201
& 53.5 \\

\rowcolor{cyan!10}
Shared Coord.
& \textbf{66,049}
& \textbf{0.1226}
& \textbf{{69.5}}
& \textbf{{0.397}}
& \textbf{54.8}\\
\bottomrule
\end{tabular}
\end{table}

\noindent\textbf{Part Consistency and Diversity.}
We further analyze whether the learned latent parts exhibit meaningful structural specialization. As shown in Table~\ref{tab:part_analysis}, the same latent part across different instances achieves a higher cosine similarity than different parts (0.595 vs. 0.459), yielding a similarity gap of 0.136 and indicating consistent part-specific representations across instances. Meanwhile, only 1.57\% of part pairs collapse to the same temporal position, while 24.38\% fall within 5\% of the local search interval. The mean normalized pairwise distance reaches 0.373, showing that different latent parts are generally instantiated at distinct temporal regions rather than concentrating on the same discriminative location.

% \begin{table}[ht]
% \centering
% \small
% \caption{Cross-instance consistency of the learned latent parts on THUMOS14. 
% Cosine similarity is computed between aligned part features from different instances of the same action class.}
% \label{tab:part_consistency}
% \begin{tabular}{lc}
% \toprule
% Metric & Cosine Similarity \\
% \midrule
% Same Part $\uparrow$       & 0.595 \\
% Different Parts $\downarrow$ & 0.459 \\
% Similarity Gap $\uparrow$  & +0.136 \\
% \bottomrule
% \end{tabular}
% \end{table}

\begin{table}[ht]
\centering
\small
\caption{Semantic consistency and temporal diversity of latent parts.}
\label{tab:part_analysis}
\setlength{\tabcolsep}{5pt}
\begin{tabular}{cc}
\toprule
Semantic Consistency & Temporal Diversity \\
\midrule
Same: 0.595 $\uparrow$ & Coll.: 1.57\% $\downarrow$ \\
Diff.: 0.459 $\downarrow$ & $CR_{0.05}$: 24.38\% $\downarrow$ \\
Gap: +0.136 $\uparrow$ & Dist.: 0.373 $\uparrow$ \\
\bottomrule
\end{tabular}
\end{table}

\noindent\textbf{Anchor Consistency Analysis.}
Table~\ref{tab:anchor_instance_consistency} shows that 95.90\% of the mined anchors belong to the target action class, and 90.45\% are located within the current action instance. Only 5.45\% come from other same-class instances, while different-class and background anchors account for 1.11\% and 2.99\%, respectively. These results indicate that the mined anchors are both class-consistent and largely instance-specific, while a small ambiguity remains among repeated same-class instances.

  \begin{table}[ht]
  \centering
  \small
  \caption{Anchor Consistency Analysis.}
  \label{tab:anchor_instance_consistency}
  \begin{tabular}{lc}
  \toprule
  Anchor Category & Fraction (\%) \\
  \midrule
  Current action instance & 90.45 \\
  Other same-class instance & 5.45 \\
  Other different-class instance & 1.11 \\
  Background & 2.99 \\
  \midrule
  Target action class & 95.90 \\
  Other classes / background & 4.10 \\
  \bottomrule
  \end{tabular}
  \end{table}

\noindent\textbf{Ablation study of DPA.}
We conduct ablations of anchor construction and deformation modeling, as shown in Table~\ref{tab:dpa_mechanism_ablation}. 
Replacing the mined anchors with uniform or random anchors decreases the AVG to 47.8 and 45.4, confirming the importance of informative instance-level anchor mining. 
Removing deformation modeling reduces the AVG to 52.2, confirming the contribution of deformation-aware matching to adapting latent parts to temporal variations.

\begin{table}[ht]
  \centering
  \caption{
  Ablation study of anchor construction and deformation modeling in DPA.}
  \label{tab:dpa_mechanism_ablation}
  \resizebox{\linewidth}{!}{
  \small
  \begin{tabular}{l|c|ccc|c}
    \toprule
    \multirow{2}{*}{Anchor}
    & \multirow{2}{*}{Deformation}
    & \multicolumn{3}{c|}{mAP@tIoU (\%)}
    & \multirow{2}{*}{\shortstack{AVG\\$[0.3{:}0.1{:}0.7]$}} \\
    % \cmidrule(lr){3-5}
    & & 0.3 & 0.5 & 0.7 & \\
    \midrule
      Mining
    & --
    & 73.9
    & 54.1
    & 23.8
    & 52.2 \\
    \rowcolor{cyan!10}
         Mining
    & \cmark
    & \textbf{{75.6}}
    & \textbf{{57.4}}
    & \textbf{{27.8}}
    & \textbf{{54.3}} \\

    % Response only
    % & \cmark
    % & 73.68
    % & 54.96
    % & 21.35
    % & 50.74 \\

    % Annotation only
    % & \cmark
    % & 72.50
    % & 53.52
    % & 22.09
    % & 50.32 \\

    Uniform
    & \cmark
    & 68.4
    & 50.9
    & 21.8
    & 47.8 \\

    Random
    & \cmark
    & 68.3
    & 47.1
    & 19.1
    & 45.4 \\

    \bottomrule
  \end{tabular}}
\end{table}

\noindent\textbf{Cross-instance Structural Consistency.}
Table~\ref{tab:part_structure_consistency} evaluates the consistency of GT-normalized part configurations across action instances. Compared with DPA, DPA+RSO yields a higher Inside rate and lower position, gap, and span dispersions, showing that RSO promotes more stable relative temporal organization among aligned parts.

\begin{table}[ht]
\centering
\small
\setlength{\tabcolsep}{3.0pt}
\caption{Cross-instance consistency of GT-normalized part configurations on THUMOS14.}
\label{tab:part_structure_consistency}
\begin{tabular}{lcccc}
\toprule
Method
& Inside $\uparrow$
& Pos. Disp. $\downarrow$
& Gap Disp. $\downarrow$
& Span Disp. $\downarrow$ \\
\midrule
DPA
& 87.6
& 0.1842
& 0.1425
& 0.1693 \\

% \rowcolor{cyan!10}
DPA + RSO
& 90.8
& 0.1517
& 0.1098
& 0.1326 \\
\bottomrule
\end{tabular}
\end{table}

\begin{figure*}[!t]
\centering
	\includegraphics[width=0.98\textwidth]{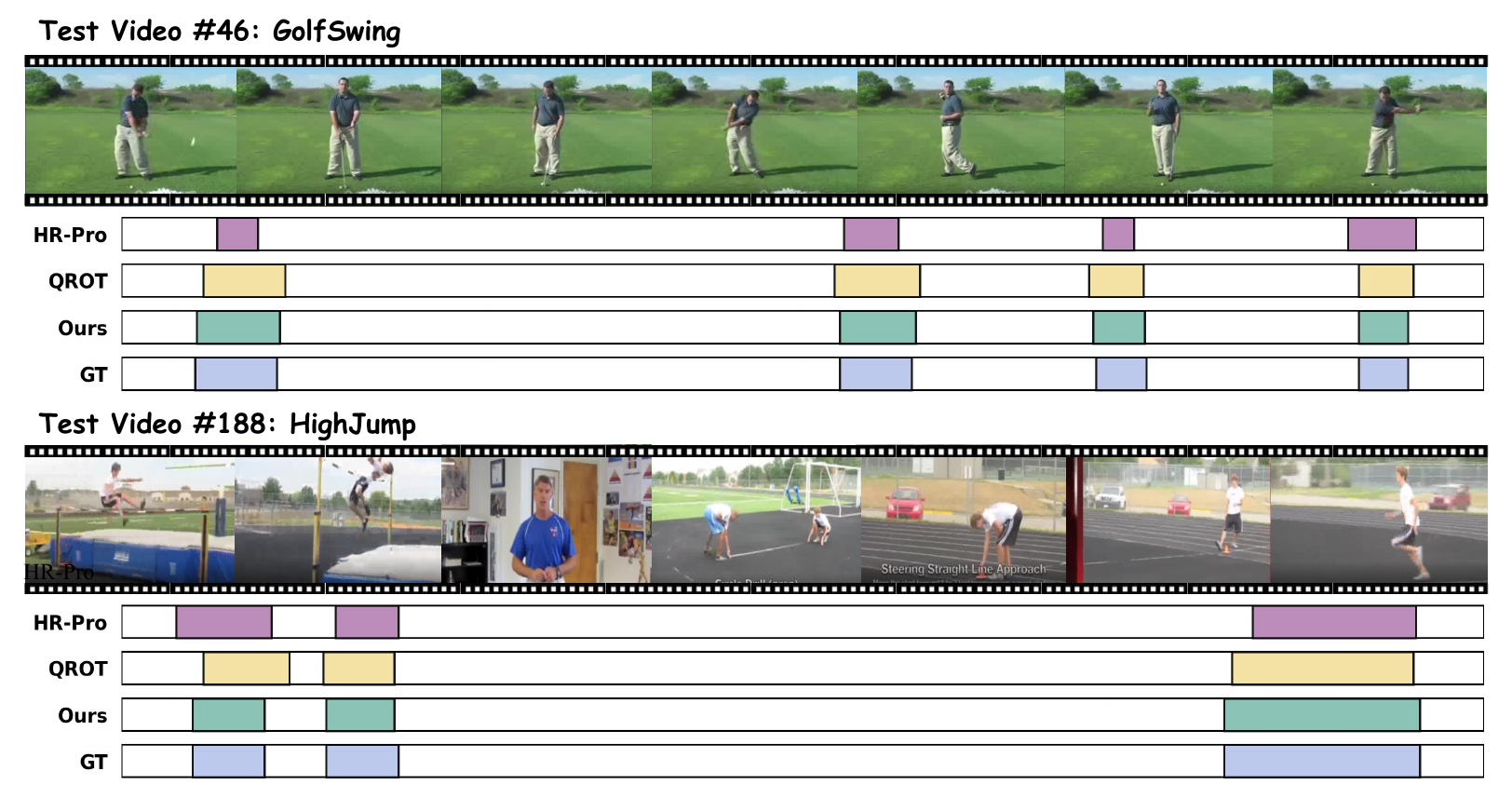}
	\caption{Qualitative comparison of our proposed method with QROT~\cite{liu2025boosting} and HR-Pro~\cite{zhang2024hr} on THUMOS14~\cite{jiang2014thumos}. We provide two test videos with action classes of GolfSwing and HighJump, respectively. The first row of each video is the input video, whereas the following three rows show the predictions from HR-Pro, QROT and Ours. The bottom row shows the ground-truth intervals. It can be seen that our detection results show higher IoUs with the ground truth.}
  \label{keshihua}
\end{figure*}

\noindent
\textbf{Efficiency comparison.} As shown in Table~\ref{tab:efficiency_qrot_full}, DeformStruct introduces only modest additional computational overhead during training, while maintaining comparable overall efficiency.

\begin{table}[ht]
\centering
\small
\setlength{\tabcolsep}{2.8pt}
\caption{
Training-time efficiency comparison.
}
\label{tab:efficiency_qrot_full}
% \resizebox{!}{\linewidth}{

\begin{tabular}{lcccc}
\toprule
Method & Params. & FLOPs & Train Runtime & Memory \\
       & (M)     & (G)   & (ms)    & (MB) \\

\midrule
\rowcolor{green!10}
QROT & 18.717 & 10.04 & 30.46 & 797.53 \\
\rowcolor{cyan!10}
DeformStruct w/o RSO & 18.930 & 10.04 & 31.10 & 798.04 \\
\rowcolor{cyan!10}
DeformStruct & 18.996& 10.04 & 33.53 & 798.29 \\

\bottomrule
\end{tabular}
\end{table}

% \begin{table}[t]
% \centering
% \small
% \caption{Inference efficiency of the completed DPA and DPA+RSO models. Measurements use an RTX 3090, batch size 1, 20 warm-up iterations, and three rounds of 100 test videos.}
% \label{tab:efficiency_qrot_full}
% \resizebox{\linewidth}{!}{%
% \begin{tabular}{lrrrrr}
% \toprule
% Method & Params. (M) & Extra RSO (M) & FLOPs (G)$^\dagger$ & Runtime (ms/video) & Memory (MiB) \\
% \midrule
% DPA & 18.930 & 0.000 & 10.04 & 31.10 & 798.04 \\
% DPA + RSO & 18.996 & 0.066 & 10.04 & 33.53 & 798.29 \\
% \bottomrule
% \end{tabular}}
% \par\smallskip
% \begin{minipage}{\linewidth}
% \footnotesize
% $^\dagger$ FLOPs are the supported-operation count from the PyTorch profiler on the first test video, excluding unsupported operations and the feature extractor. Runtime includes the model forward pass and prediction postprocessing, excluding data loading, host-to-device transfer, and evaluator soft-NMS. Memory is peak allocated inference memory. Measurements were made on a shared GPU. RSO is not called during inference.
% \end{minipage}
% \end{table}

\noindent
\textbf{Robustness analysis under different point distributions.}
We further evaluate DeformStruct under different point-annotation distributions. As shown in Table~\ref{tab:point_distribution_robustness}, the model maintains competitive overall performance under random, center-biased, and discriminative-frame annotations, although the performance varies across different annotation distributions and tIoU thresholds. In particular, center-biased annotations achieve higher average performance, while random and discriminative-frame annotations remain reasonably close to the original dataset setting. These results indicate that DeformStruct can accommodate different point-annotation distributions without relying on a specific annotation pattern.

% \begin{figure}[h]
% \centering
% \begin{minipage}{0.48\linewidth}
%     \centering
%     \includegraphics[width=\linewidth]{figure/point_position_histogram.pdf}
% \end{minipage}
% \hfill
% \begin{minipage}{0.48\linewidth}
%     \centering
%     \includegraphics[width=\linewidth]{figure/point_position_regions.pdf}
% \end{minipage}
% \caption{
% \textcolor{blue}{
% Distribution analysis of point annotations within ground-truth action instances. 
% The left figure shows the normalized temporal distribution of all point annotations across datasets. The right figure reports the proportions of point annotations falling into the early, middle, and late regions for each dataset. 
% }}
% \label{fig:point_position_distribution}

% \end{figure}

% \begin{figure}[h]
% \centering
% \includegraphics[width=\linewidth]{figure/point_strategy_histograms.pdf}
% \caption{
% \textcolor{blue}{
% Visualization of simulated point annotation distributions. 
% We consider four strategies: random, center-biased, discriminative-frame, and boundary-biased points. 
% Dashed vertical lines denote the boundaries between early, middle, and late regions.
% }}
% \label{fig:point_distribution_strategies}
% \end{figure}

\begin{table}[h]
\centering
\small
\caption{Robustness analysis under different point annotation distributions. All variants use DeformStruct and differ only in the point-annotation distribution.}
\label{tab:point_distribution_robustness}
% \resizebox*{\linewidth}{!}{
\begin{tabular}{l|ccc|c}
\toprule
 \multirow{2}{*}{Setting} & \multicolumn{3}{c|}{mAP@tIoU (\%)} & \multicolumn{1}{c}{AVG} \\ 
 &0.3&0.5&0.7 &  [0.3:0.1:0.7] \\ 
\midrule
\rowcolor{cyan!10}
Dataset-distribution  & \textbf{76.4} &57.9& 28.0& 54.8 \\
Random Point& 71.1 & 56.7 & 28.3 & 53.4\\
Center-biased Point & 71.5 & \textbf{61.6} & \textbf{30.7}  &\textbf{56.5}\\
Discriminative-frame Point & 76.0 & 58.3 & 26.3 &54.7 \\
% Boundary-biased Point & 69.5 & 48.5 & 17.5 & 46.4\\

\bottomrule
\end{tabular}
% }
\end{table}

% \noindent\textcolor{blue}{\textbf{Analysis of the clusters and temporal regions. }}
% \textcolor{blue}{
% Fig.~\ref{fig:qr_cluster_temporal_analysis} analyzes the temporal distributions of Query Reformation feature clusters after mapping them back to ground-truth action timelines. 
% Each row represents one of the top 30 clusters with the most assigned point annotations. 
% The thin line denotes the min--max range, the thick bar denotes the interquartile range, and the markers indicate the mean and median positions. 
% Although K-means is performed only in the feature space, different feature clusters exhibit diverse temporal coverage when mapped
% back to the normalized action timeline.
% }
% \input{tables-and-figures/cluster_distru}

\begin{table*}[t]
	\caption{Hyperparameter sensitivity analysis.}
	\label{tab:hyperparameter}

	\begin{minipage}{0.32\linewidth}
		\centering
		\captionsetup[subfloat]{position=top,font=footnotesize}
		\subfloat[Sensitivity analysis on the suppression radius hyperparameter ($\delta$) for SAM.]{
			%\label{tab:hyperparameter-beta}
	 		\begin{tabular}{c|ccc|c}
	 			\toprule
	 			\multirow{2}{*}{$\delta$} & \multicolumn{3}{c|}{mAP@tIoU (\%)} & \multicolumn{1}{c}{AVG}  \\
	 			& 0.3 & 0.5 & 0.7 & 0.3:0.1:0.7 \\
	 			\midrule
	 		
	 			0& 75.3 & 56.9 & 23.5 & 53.4 \\
				   \rowcolor{cyan!10}  $0.01$  & \textbf{76.4} & \textbf{57.9} & \textbf{28.0}& \textbf{54.8} \\
                $0.05$ &  74.3 & 56.3 & 25.7 & 53.6   \\
				     $0.1$ & 74.6 & 54.3 & 24.5 & 52.0 \\
	 	        \bottomrule
	 		\end{tabular}
}
	\end{minipage}
	\hfil
	\begin{minipage}{0.32\linewidth}
		\centering
		\captionsetup[subfloat]{position=top,font=footnotesize}
		\subfloat[Sensitivity analysis on the deformation regularization weight hyperparameter ($\lambda_{\mathrm{deform}}$) for DPA.]{
	 		\begin{tabular}{c|ccc|c}
	 			\toprule
	 			\multirow{2}{*}{$\lambda_{deform}$} & \multicolumn{3}{c|}{mAP@tIoU (\%)} & \multicolumn{1}{c}{AVG}  \\
	 			& 0.3 & 0.5 & 0.7 & 0.3:0.1:0.7 \\
	 			\midrule
				0.5 & 75.5 & 53.4 & 24.1 & 51.6 \\ 
				\rowcolor{cyan!10}1 & \textbf{76.4} & \textbf{57.9} & \textbf{28.0}& \textbf{54.8} \\
				2 & 75.3 & 57.5 & 23.3 & 53.5 \\
				4 & 74.6  & 55.6  & 23.3  & 52.3 \\
	 	        \bottomrule
	 		\end{tabular}
		}
	\end{minipage}
    \hfil
    \begin{minipage}{0.32\linewidth}
		\centering
		\captionsetup[subfloat]{position=top,font=footnotesize}
		\subfloat[Sensitivity analysis on the cluster number hyperparameter ($q$) for QR.]{
			%\label{tab:hyperparameter-alpha}
	 		\begin{tabular}{c|ccc|c}
	 		\toprule
	 			\multirow{2}{*}{$q$} & \multicolumn{3}{c|}{mAP@tIoU (\%)} & \multicolumn{1}{c}{AVG}  \\
	 			& 0.3 & 0.5 & 0.7 & 0.3:0.1:0.7 \\
	 			\midrule
				0 & 74.6 & 55.5 & 25.4 & 52.7 \\
				2 &73.7 & 55.4 & 22.5 & 51.8\\
                \rowcolor{cyan!10}3  & \textbf{76.4} & \textbf{57.9} & \textbf{28.0}& \textbf{54.8} \\
				5 & 73.6 & 55.2 & 22.4 & 51.7 \\
	 	       \bottomrule
	 		\end{tabular}
		}
	\end{minipage}

    \vspace{10pt}

    \begin{minipage}{0.32\linewidth}
		\centering
		\captionsetup[subfloat]{position=top,font=footnotesize}
		\subfloat[Sensitivity analysis on the number of latent parts hyperparameter ($N_{part}$) for DPA.]{
			%\label{tab:hyperparameter-alpha}
	 		\begin{tabular}{c|ccc|c}
	 			\toprule
	 			\multirow{2}{*}{$N_{part}$} & \multicolumn{3}{c|}{mAP@tIoU (\%)} & \multicolumn{1}{c}{AVG}  \\
	 			& 0.3 & 0.5 & 0.7 & 0.3:0.1:0.7 \\
	 			\midrule
				2 & 75.1 &  55.9 &   23.5 & 52.5  \\
	 		 \rowcolor{cyan!10}3 & \textbf{76.4} & \textbf{57.9} & \textbf{28.0}& \textbf{54.8} \\
	 			4 & 75.5 &  57.3 & 25.2  & 53.0  \\
	 			5 & 75.2 &  56.1 &  26.3 &  53.5 \\
	 	        \bottomrule
	 		\end{tabular}
		}
	\end{minipage}
	\hfil
	\begin{minipage}{0.32\linewidth}
		\centering
		\captionsetup[subfloat]{position=top,font=footnotesize}
		\subfloat[Sensitivity analysis on the mining size hyperparameter ($\text{top-}{K_a}$) for SAM.]{
	 		\begin{tabular}{c|ccc|c}
	 			\toprule
	 			\multirow{2}{*}{${K_a}$} & \multicolumn{3}{c|}{mAP@tIoU (\%)} & \multicolumn{1}{c}{AVG}  \\
	 			& 0.3 & 0.5 & 0.7 & 0.3:0.1:0.7 \\
	 			\midrule
				3 & 73.2 & 54.6 & 24.5 & 52.3 \\
				5 & 75.5 & 57.8 & 26.1 & 53.7 \\
				\rowcolor{cyan!10}8 & \textbf{76.4} & \textbf{57.9} & \textbf{28.0}& \textbf{54.8} \\
				10 & 75.8 & 56.4 & 26.0 & 53.8 \\
	 	        \bottomrule
	 		\end{tabular}
		}
	\end{minipage}
	\hfil
	\begin{minipage}{0.32\linewidth}
		\centering
		\captionsetup[subfloat]{position=top,font=footnotesize}
		\subfloat[Sensitivity analysis on the balancing weight hyperparameter ($\lambda_{OT}$) for OT.]{
			%\label{tab:hyperparameter-beta}
	 		\begin{tabular}{c|ccc|c}
	 			\toprule
	 			\multirow{2}{*}{$\lambda_{OT}$} & \multicolumn{3}{c|}{mAP@tIoU (\%)} & \multicolumn{1}{c}{AVG}  \\
	 			& 0.3 & 0.5 & 0.7 & 0.3:0.1:0.7 \\
	 			\midrule
	 			0.1 & 73.5 & 53.0 & 20.1 & 50.3 \\
				\rowcolor{cyan!10}0.5 & \textbf{76.4} & \textbf{57.9} & \textbf{28.0}& \textbf{54.8} \\
				1 & 75.1 & 56.9 & 22.7 & 53.0\\
				2& 74.3 & 53.7 & 20.6 & 50.7 \\
	 	        \bottomrule
	 		\end{tabular}
		}
	\end{minipage}
\end{table*}

\noindent
\textbf{Qualitative visualization of the DPA module.} In Fig.~\ref{part_visual}, we provide visualizations of the score distributions of different parts in the DeformStruct module. The colored markers indicate the locations of anchors matched with latent action parts. The bars at the bottom indicate the ground-truth segments, while the three curves correspond to the response scores of different latent part classifiers. Notably, the mined anchors align well with the ground-truth action intervals and capture complementary regions distributed over time.

\noindent\textbf{Visualization of Part Deformation.}
Fig.~\ref{fig:instance_deformations} visualizes the deformation performed by DPA. The matched anchors are not directly used as the final part locations; instead, each latent part can shift according to its response and deformation cost.
As shown in the examples, different parts exhibit different deformation directions and magnitudes, while some remain close to their anchors when further adjustment is unnecessary.
This qualitatively shows that DPA adaptively refines the mined anchors rather than simply inheriting their temporal positions.

\noindent\textbf{Visualization of Learned Structural Coordinates.}
Fig.~\ref{fig: rso} compares the normalized aligned-part coordinates ${p_i}$ with the structural coordinates ${q_i}$ predicted from aligned part features. After mean-centering, the learned structural coordinates exhibit configurations consistent with the anchor structure while allowing flexible global shifts. The pairwise difference plots further show that RSO preserves the relative ordering and spacing among parts. These results demonstrate that RSO captures the underlying relational structure without forcing the learned coordinates to exactly match the temporal anchors.

% \noindent\textbf{Boundary-error analysis.}
% As shown in Fig.~\ref{fig:boundary_error_combined1}, DeformStruct achieves lower overall boundary errors than QROT. 
% The normalized start, end, center, and duration errors are all reduced, indicating improved boundary robustness. 
% The improvements are more evident for short- and medium-duration actions, whose boundaries are harder to infer under point-level supervision. 
% When predictions are grouped by matched temporal IoU, DeformStruct mainly reduces errors in low- and moderate-IoU regions, while the gap becomes smaller in the high-IoU region. 
% This explains the relatively lower gain at IoU=0.7: DeformStruct mainly improves action-region coverage by aggregating complementary action-related evidence, rather than explicitly refining exact boundaries.
% This behavior is consistent with the possibility that DPA may include nearby transition or contextual snippets when aggregating action-related evidence.
% Such expansion is less harmful at moderate IoU thresholds but is penalized under strict high-IoU evaluation. 
% Therefore, DeformStruct improves overall boundary robustness, while its advantage becomes less pronounced at IoU=0.7.

\noindent
\textbf{Sensitivity analysis of the hyperparameters.}
We further study the sensitivity of DeformStruct to its main hyperparameters in Table~\ref{tab:hyperparameter}, including the suppression radius $\delta$, deformation regularization weight $\lambda_{deform}$, cluster number $q$, number of latent parts $N_{part}$, mining size top-$K_a$, and OT balancing weight $\lambda_{OT}$. The results show that the model achieves relatively stable performance within moderate ranges of these hyperparameters, whereas excessively small or large values may lead to performance degradation.
% The adopted configuration yields consistently strong overall performance across the evaluated tIoU thresholds.

\subsection{Qualitative Results}
To further validate the effectiveness of the proposed method, we provide some qualitative results using HR-Pro~\cite{zhang2024hr}, QROT~\cite{liu2025boosting} and our model on test videos \#46 and \#188 in THUMOS14~\cite{jiang2014thumos} for comparison. Several representative results are visualized in Fig.~\ref{keshihua}, illustrating the improved localization quality of our approach.
% }

\section{Conclusion}
\label{sec:conclusion}
In this work, we propose DeformStruct, an end-to-end framework for point-supervised temporal action localization that focuses on instance-level structural modeling under sparse supervision. DeformStruct progressively discovers complementary action-related evidence, instantiates latent parts through deformation-aware alignment, and explicitly models their relative temporal organization. Semantic query construction and OT-based pseudo-segment generation are retained to provide point-level semantic priors and segment-level supervision, respectively. Extensive experiments on THUMOS14, ActivityNet-v1.3, GTEA, and BEOID demonstrate that DeformStruct consistently outperforms existing point-supervised methods, while detailed ablations and structural analyses further validate the effectiveness of part discovery, deformable alignment, and relational organization.
% \endgroup

% \section{Limitations.}
% \textcolor{blue}{
% Although the proposed method improves action completeness and semantic coverage under point-level supervision, it still has several limitations. 
% First, since ground-truth temporal boundaries are unavailable, DPA cannot directly optimize precise start and end boundaries, and may still suffer from boundary drift under strict IoU thresholds. 
% Second, our method relies on pre-extracted I3D features, so its performance may be limited by the representation quality and generalization ability of the feature extractor. 
% Exploring stronger or end-to-end trainable video representations is a promising future direction. 
% Finally, temporal action localization may be used in surveillance-like scenarios, raising concerns about privacy, fairness, and potential misuse. 
% Thus, practical deployment should consider privacy protection, transparent data usage, and necessary human oversight.
% }

% \section{References Section}
% You can use a bibliography generated by BibTeX as a .bbl file.
%  BibTeX documentation can be easily obtained at:
%  http://mirror.ctan.org/biblio/bibtex/contrib/doc/
%  The IEEEtran BibTeX style support page is:
%  http://www.michaelshell.org/tex/ieeetran/bibtex/
 

% \fi

\bibliographystyle{IEEEtran}
\bibliography{egbib}

% \newpage



% % \bf{If you include a photo:}\vspace{-33pt}
% \end{IEEEbiography}



%\vspace{-10mm}
\begin{IEEEbiography}[{\includegraphics[width=1in,height=1.25in,clip,keepaspectratio]{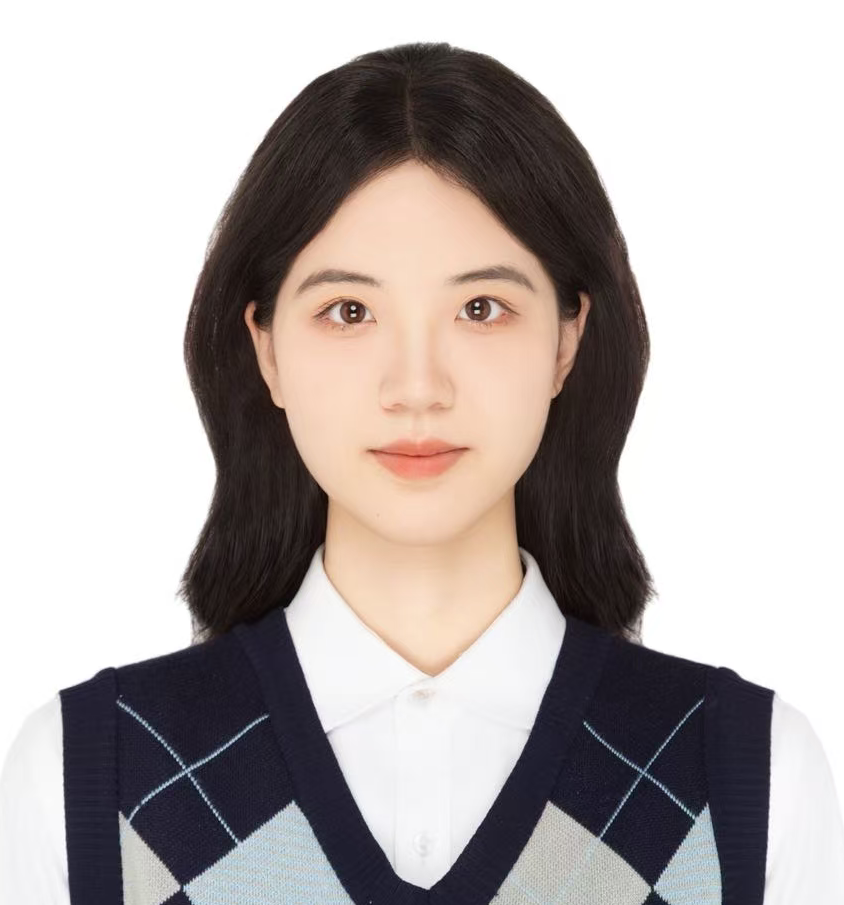}}]{Mengnan Liu} received the B.S. degree in Artificial Intelligence from Shandong University, Jinan, China, in 2022. She is currently pursuing the Ph.D. degree at the Institute of Artificial Intelligence and Robotics, Xi’an Jiaotong University. Her research interests include computer vision, multimodal video understanding and weakly-supervised learning.
\end{IEEEbiography}
%\vspace{-10mm}

\begin{IEEEbiography}[{\includegraphics[width=1in,height=1.25in,clip,keepaspectratio]{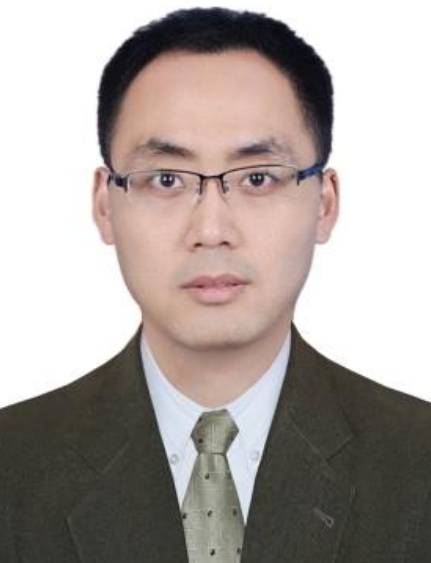}}]{Sanping Zhou}
(Member, IEEE) received his Ph.D. degree from Xi'an Jiaotong University, Xi'an, China, in 2020. From 2018 to 2019, he was a Visiting Ph.D. Student with the Robotics Institute, Carnegie Mellon University.  He is currently a Professor with the Institute of Artificial Intelligence and Robotics at Xi'an Jiaotong University.  His research interests include machine learning, computer vision and embodied intelligence, with a focus on meta-learning, multi-task learning, object detection, multi-target tracking, trajectory prediction, medical image segmentation, visual navigation, and visual grasping.
\end{IEEEbiography}
%\vspace{-10mm}

\begin{IEEEbiography}[{\includegraphics[width=1in,height=1.25in,clip,keepaspectratio]{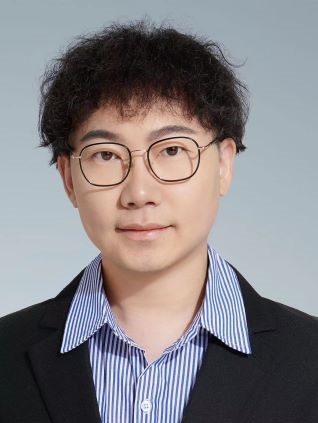}}]{Kun Xia}
received the Ph.D. degree in Control Science and Engineering from Xi'an Jiaotong University, Xi'an, China, in 2024. 
From 2022 to 2023, he was a visiting Ph.D. student at the University of Illinois Chicago, IL, USA. He is currently an Assistant Professor with the School of Computer Science and Technology, Xi'an Jiaotong University, Xi'an, China. 
His research interests include computer vision, image/video processing, analysis and understanding.
\end{IEEEbiography}
%\vspace{-10mm}

\begin{IEEEbiography}
[{\includegraphics[width=1in,height=1.25in,clip,keepaspectratio]{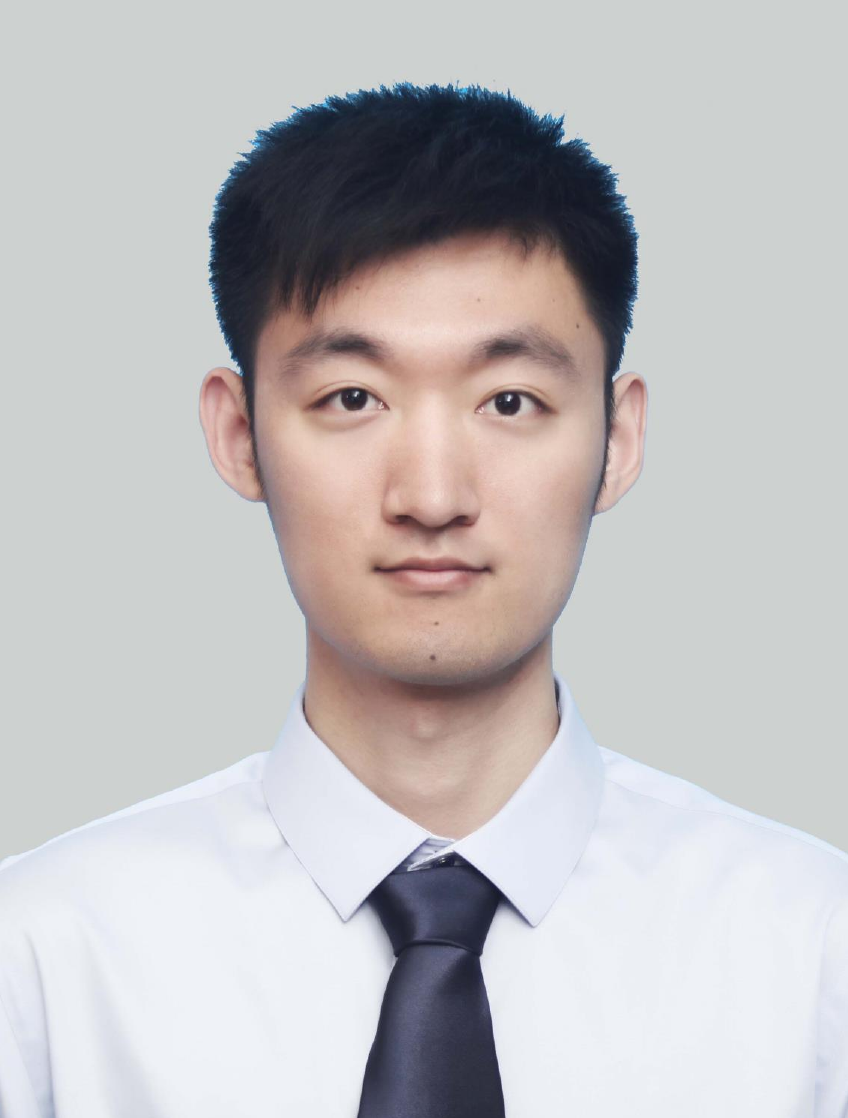}}]
{Yonghao Dong}
 received the Ph.D. degree in Electronic Information from Xi’an Jiaotong University, Xi’an, China, in 2025. He received the B.S. degree from East China University of Science and Technology, Shanghai, China, in 2018, and the M.S. degree from the University of Illinois at Chicago, Chicago, IL, USA, in 2019. He is currently an Assistant Professor with the School of Software Engineering, Xi’an Jiaotong University, Xi’an, China. His research interests include computer vision, image/video processing and autonomous driving.
\end{IEEEbiography}
%\vspace{-10mm}

\begin{IEEEbiography}
[{\includegraphics[width=1in,height=1.25in,clip,keepaspectratio]{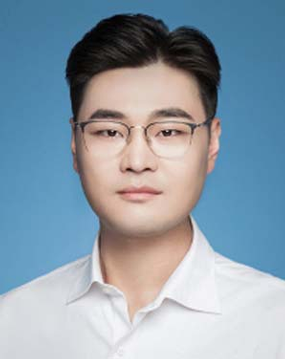}}]{Jie Wang} (Senior Member, IEEE) received the BSc degree in electronic information science and technology from the University of Science and Technology of China, Hefei, China, in 2005, and the PhD degree in computational science from the Florida State University, Tallahassee, FL, in 2011. He is currently a professor with the Department of Electronic Engineering and Information Science, University of Science and Technology of China. His research interests include reinforcement learning, knowledge graph, large-scale optimization, deep learning, etc.
\end{IEEEbiography}
% \begin{IEEEbiography}[{\includegraphics[width=1in,height=1.25in,clip,keepaspectratio]{images/author/Nanning-Zheng}}]{Nanning Zheng}
% (Fellow, IEEE) graduated from the Department of Electrical Engineering of Xi'an
% Jiaotong University (XJTU), Xi'an, China, in 1975, received the M.E. degree in
% Information and Control Engineering from Xi'an Jiaotong University, Xi'an,
% China, in 1981, and the Ph.D. degree in electrical engineering from Keio
% University, Keio, Japan, in 1985. He is currently a Professor and the Director
% with the Institute of Artificial Intelligence and Robotics of Xi'an Jiaotong
% University, Xi'an, China. His research interests include computer vision,
% pattern recognition, computational intelligence, and hardware implementation of
% intelligent systems. Since 2000, he has been the Chinese representative on the
% Governing Board of the International Association for Pattern Recognition. He
% became a member of the Chinese Academy of Engineering in 1999.
% \end{IEEEbiography}

\begin{IEEEbiography}[{\includegraphics[width=1in,height=1.25in,clip,keepaspectratio]{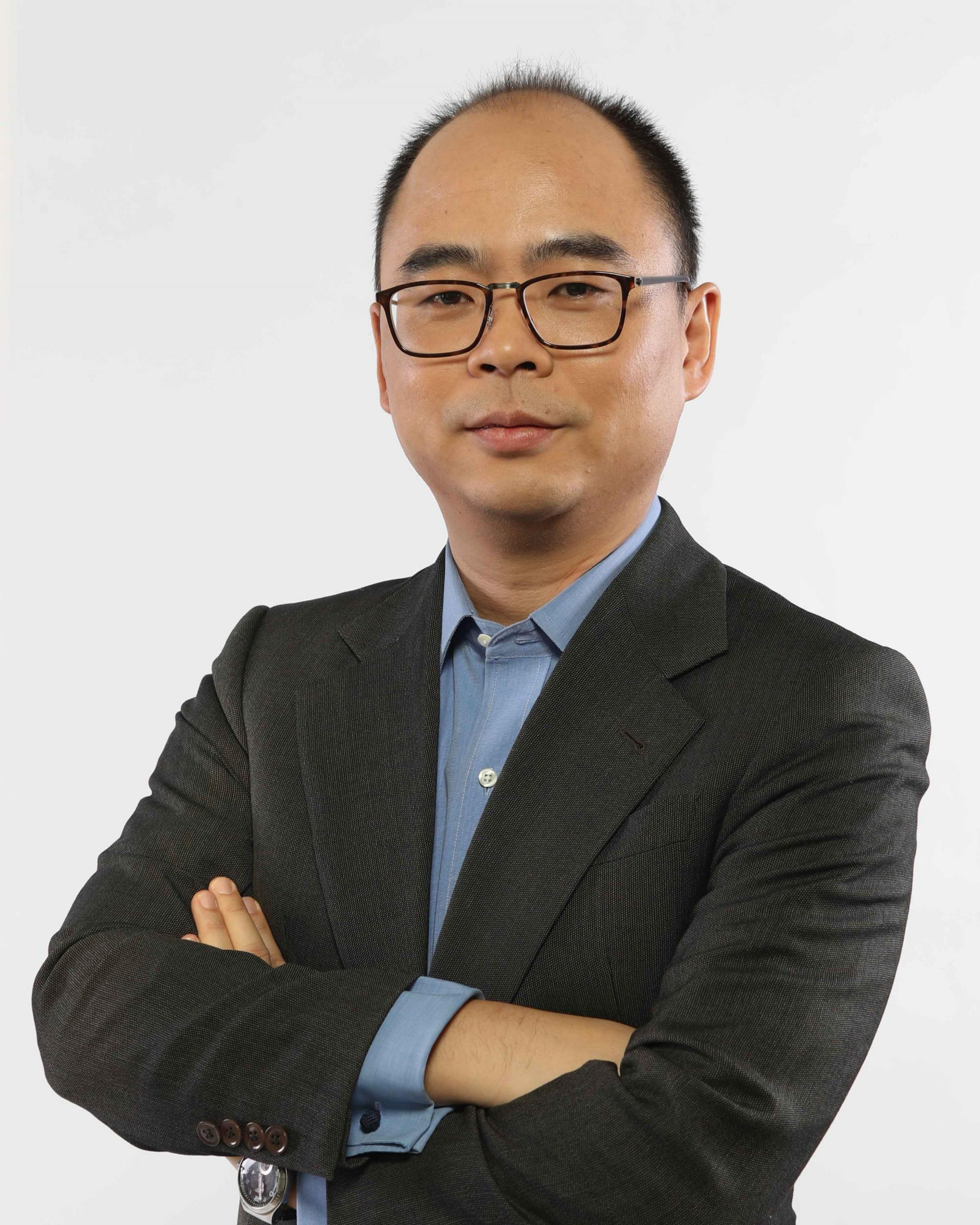}}]{Gang Hua} (Fellow, IEEE) received the B.S. and M.S. degrees in Automatic Control Engineering from Xi’an Jiaotong University (XJTU), Xi’an, China, in 1999 and 2002, respectively. He received the Ph.D. degree in Electrical Engineering and Computer Science at Northwestern University, Evanston, Illinois, USA, in 2006. He is currently the  Director of Applied Science at Amazon Alexa AI. He is an associate editor of TPAMI and MVA. He is a general chair of ICCV’2025, a program chair of CVPR’2019\&2022, and an area chair of CVPR’2015\&2017 and ICCV’2011\&2017. He is the author of more than 200 peer-reviewed publications in prestigious international journals and conferences. He holds 19 US patents and has 15 more US patents pending. He is the recipient of the 2015 IAPR Young Biometrics Investigator Award. He is an IEEE Fellow, an IAPR Fellow, and an ACM Distinguished Scientist.
\end{IEEEbiography}

\end{document}